%% file: neurips_2026.tex
\documentclass{article}

\PassOptionsToPackage{authoryear}{natbib}
\usepackage[main,final]{neurips_2026}

\usepackage[utf8]{inputenc}
\usepackage[T1]{fontenc}
\usepackage{amsmath,amssymb}
\usepackage{graphicx}
\usepackage{booktabs,array,multirow,tabularx,colortbl}
\usepackage{caption}
\usepackage{float}
\usepackage{listings}
\usepackage{anyfontsize}
\usepackage{microtype}
\usepackage{xcolor}
\usepackage{hyperref}
\usepackage{url}

\definecolor{linkblue}{HTML}{68ACE5}
\definecolor{citationgray}{gray}{0.5}
\definecolor{Gray}{gray}{0.93}
\hypersetup{colorlinks=true,linkcolor=linkblue,citecolor=citationgray,urlcolor=linkblue}

\newcommand{\abbr}{\textsc{AgentOdyssey}}

\definecolor{darkbrown}{RGB}{110,60,20}
\definecolor{darkroyalblue}{RGB}{40,70,160}
\definecolor{darkmagenta}{RGB}{150,20,110}
\definecolor{darkcyan}{RGB}{0,110,110}
\definecolor{darkolivegold}{RGB}{140,120,20}

\newcommand{\exploration}{\textcolor{darkbrown}{exploration}}
\newcommand{\episodicmemory}{\textcolor{darkroyalblue}{episodic memory}}
\newcommand{\worldknowledge}{\textcolor{darkmagenta}{world knowledge acquisition}}
\newcommand{\skilllearning}{\textcolor{darkcyan}{skill learning}}
\newcommand{\longhorizonplanning}{\textcolor{darkolivegold}{long-horizon planning}}

\newcommand{\Exploration}{\textcolor{darkbrown}{Exploration}}
\newcommand{\EpisodicMemory}{\textcolor{darkroyalblue}{Episodic Memory}}
\newcommand{\WorldKnowledge}{\textcolor{darkmagenta}{World Knowledge Acquisition}}
\newcommand{\SkillLearning}{\textcolor{darkcyan}{Skill Learning}}
\newcommand{\LongHorizonPlanning}{\textcolor{darkolivegold}{Long-Horizon Planning}}

\newcommand{\graymidrule}{\arrayrulecolor{gray}\midrule\arrayrulecolor{black}}

\newcommand{\heart}{\ensuremath\heartsuit}

\definecolor{codegreen}{rgb}{0,0.6,0}
\definecolor{codegray}{rgb}{0.5,0.5,0.5}
\definecolor{codepurple}{rgb}{0.58,0,0.82}
\definecolor{backcolour}{rgb}{0.95,0.95,0.92}

\lstdefinestyle{mycode}{
    backgroundcolor=\color{backcolour},
    commentstyle=\color{codegreen},
    keywordstyle=\color{magenta},
    stringstyle=\color{codepurple},
    basicstyle=\ttfamily\fontsize{7.5}{7.5}\selectfont,
    frame=single, 
    breakatwhitespace=false,
    breaklines=true,
    captionpos=b,
    keepspaces=true,
    showspaces=false,
    showstringspaces=false,
    showtabs=false,
    tabsize=4
}

\newcommand{\beginsupplement}{%
  \appendix
  \setcounter{table}{0}
  \renewcommand{\thetable}{A\arabic{table}}%
  \renewcommand{\theequation}{A\arabic{equation}}%
  \setcounter{figure}{0}
  \renewcommand{\thefigure}{A\arabic{figure}}%
  \setcounter{section}{0}
  \renewcommand{\thesection}{A\arabic{section}}%
}

\title{AgentOdyssey: Open-Ended Long-Horizon Text Game Generation for Test-Time Continual Learning Agents}

\author{%
  Zheyuan Zhang\thanks{Equal contribution. $^{\heart}$ Equal advising.}\hspace{0.4em}$^1$,
  Zehao Wen$^{*1}$,
  Alvin Zhang$^1$,
  Andrew Wang$^1$,
  Jianwen Xie$^2$ \\
  \textbf{Daniel Khashabi$^{\heart1}$,
  Tianmin Shu$^{\heart1}$} \\
  \normalfont{$^{1}$ Johns Hopkins University \qquad $^{2}$ Lambda} \\[3pt]
  \texttt{\href{https://agentodyssey.github.io/}{AgentOdyssey.github.io}}
}

\begin{document}
\maketitle

\begin{abstract}
For agents to learn continuously from interaction with the world at test time, they must be able to explore effectively, acquire new world knowledge and skills, retain relevant episodic experiences, and plan over long horizons. To evaluate these key abilities of test-time continual learning agents, we introduce \abbr{}, a novel evaluation framework that procedurally generates open-ended text games with rich entities, world dynamics, and long-horizon tasks. Critically, \abbr{} goes beyond the conventional machine learning assumption that learning does not occur at test time by placing agents in a continuous, long-horizon setting that interleaves learning and inference throughout deployment. We further propose a multifaceted evaluation methodology that measures not only game progress but also offers diagnostic tests on world knowledge acquisition, episodic memory, object and action exploration, action diversity, and model cost. We evaluate diverse agent paradigms across six generated games. Our experimental results reveal critical limits in agents' key abilities, as well as factors that influence their meaningful horizon. Although performance scales with stronger base models, even the top agent remains far below human performance on a demanding long-horizon game, leaving substantial headroom for improvement. Additionally, a focused study of agent test-time training finds that, although short-term memory is beneficial, agents trained with supervised fine-tuning still suffer from catastrophic forgetting, suggesting that parametric continual learning for agents remains an open challenge.
\end{abstract}

\begin{figure*}[t]
    \centering
    \includegraphics[width=\textwidth]{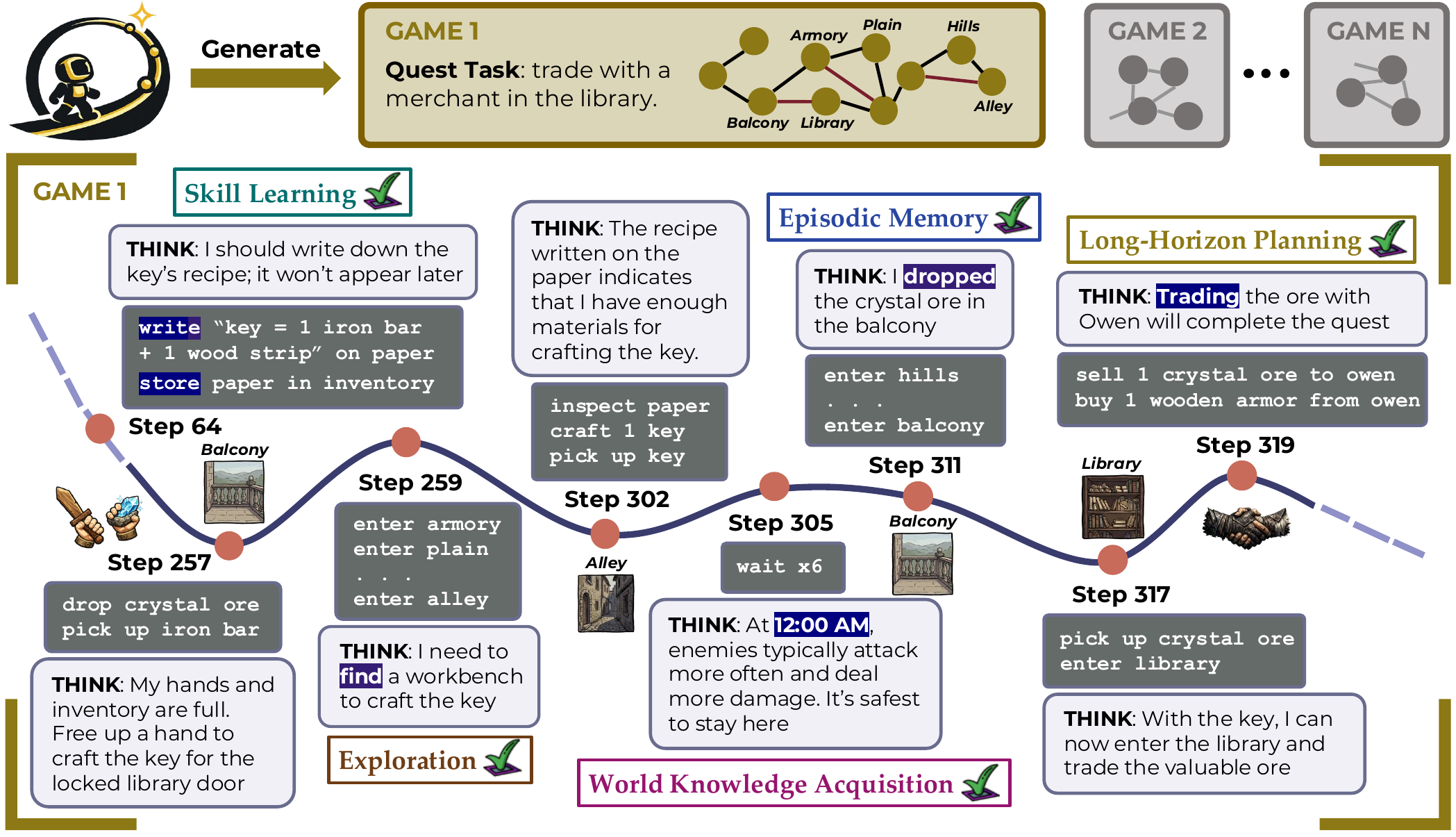}
    \caption{\textbf{Example player trajectory in an \abbr{}-generated game.} \abbr{} games test five key abilities for test-time continual learning agents: \textbf{\exploration} for finding the workbench, \textbf{\episodicmemory} for recalling the dropped trade item, \textbf{\worldknowledge} for avoiding dangerous travel times, \textbf{\skilllearning} for improving crafting via written recipes, and \textbf{\longhorizonplanning} for completing multi-step objectives such as crafting a key for trade. The trajectory shows these abilities working together over hundreds of steps to complete the quest.}
    \label{fig:teaser}
\end{figure*}

\input{sections/intro}
\input{sections/related}
\input{sections/env}
\input{sections/agent}
\input{sections/exp1}
\input{sections/exp2}

\section{Conclusion}
We introduce \abbr{}, an open-ended text game generation framework for evaluating test-time continual learning agents in long-horizon, non-resettable environments, together with multifaceted diagnostics that probe key abilities beyond task rewards. Experiments across diverse agent paradigms and six generated games show that performance scales with memory capacity and backbone reasoning ability, while revealing persistent limitations in \exploration, \episodicmemory, \worldknowledge, \skilllearning, and \longhorizonplanning. Through a focused study of agent test-time training, we further find that although short-term memory provides effective working memory, parametric test-time training still suffers from catastrophic forgetting, highlighting the importance of continual learning that both mitigates forgetting and preserves the model's plasticity to keep acquiring new knowledge when updating weights at test time. Overall, our results highlight these challenges and motivate future work on architectures and algorithms for persistent memory and scalable test-time learning.

\textbf{Limitations and Future Work.} Our environment currently supports only textual observations and a single agent. Its turn-based design assigns each action a fixed duration, simplifying real-world temporal dynamics. These limitations constrain the study of visual perception and grounding, multi-agent interaction, and world dynamics modeling. In future work, \abbr{} could be extended to generate multi-agent games with visual rendering and richer temporal dynamics. Appendix~\ref{app:generator_evaluation} discusses generation quality and qualitative failure cases.

\begin{ack}
This project was sponsored by a generous award from Amazon. We thank our colleagues and collaborators for their input on an earlier draft of this work. TS also acknowledges Lambda for its support in providing computational resources.

\end{ack}

\clearpage

\bibliography{references}

\input{sections/appendix}

\end{document}

%% file: sections/intro.tex
\section{Introduction}
\begin{table*}[t]
\setlength{\tabcolsep}{1.5pt}
\renewcommand{\arraystretch}{1.25}
\centering
\small
\caption{\textbf{Comparison of text game environments for agents.} Columns report, from left to right: \textbf{World} (open or closed), \textbf{\# E}ntities (types of entities, O = objects, A = areas, N = NPCs), \textbf{\# A}ctions (number of action types per game), \textbf{\# G}oals (number of narrative goals per game), \textbf{Cont}amination (whether evaluation content may appear in training data), \textbf{Scal}ability (ability to generate larger or varied game environments), \textbf{Stoch}astic (presence of random dynamics), \textbf{Dyn}amic World (whether the game environment evolves independently of agent actions), \textbf{Hor}izon (typical episode length in steps). We use the game environment in Voyager \citep{wang2023voyager} as a text-based Minecraft setting, where agents receive world observations in textual form and produce executable code actions. For \abbr{}, Config. means configurable during generation and applies to the number of action rules, goals, and game horizon. The six evaluated games average 20 action rule types and 19.8 main quest stages. Agent evaluations use a budget of up to 500 steps, which is not a framework limit.
}
\begin{tabular*}{\textwidth}{@{\extracolsep{\fill}}l*{9}{c}@{}}
\toprule
\textbf{Environment} 
& \textbf{World} 
& \textbf{\# E} 
& \textbf{\# A} 
& \textbf{\# G} 
& \textbf{Cont.} 
& \textbf{Scal.} 
& \textbf{Stoch.} 
& \textbf{Dyn.}
& \textbf{Hor.}
\\
\midrule
\addlinespace

\shortstack[l]{ALFWorld\\\citep{shridhar2020alfworld}}
& Close 
& O,A 
& 11
& 1 
& Yes 
& No
& No
& No
& 50
\\

\graymidrule

\shortstack[l]{Jericho\\\citep{hausknecht2020interactive}}
& Close 
& \textbf{O, A, N}
&187.2
& 1
& Yes 
& No 
& \textbf{Yes} 
& No
& 100
\\
\graymidrule

\shortstack[l]{ByteSized32\\\citep{wang2023bytesized32}}
& Close 
& O
& 9.8 
& 1 
& \textbf{No} 
& \textbf{Yes} 
& No 
& No
& 12.8 \\
\graymidrule

\shortstack[l]{Minecraft (Text)\\\citep{wang2023voyager}}
& \textbf{Open} 
& \textbf{O, A, N}
& $\sim$30 
& 0 
& Yes 
& No 
& \textbf{Yes} 
& \textbf{Yes}
& $\boldsymbol{\infty}$ \\
\graymidrule

\textbf{\abbr{}}
& \textbf{Open} 
& \textbf{O, A, N}
& Config.
& Config.
& \textbf{No} 
& \textbf{Yes}
& \textbf{Yes} 
& \textbf{Yes}
& Config. \\
\bottomrule
\end{tabular*}
\label{tab:text_games}
\end{table*}

Advances in large language models (LLMs) have demonstrated strong reasoning and problem-solving capabilities in challenging domains such as mathematics and programming \citep{hendrycks2021measuring, chen2021codex, wei2022chain, kojima2022large, roziere2023code, bai2023qwen, lightman2023let, guo2025deepseek}. Building on these successes, there have been many LLM-based agentic frameworks for sequential decision-making and embodied tasks, where LLM-based agents perceive, plan, and act in interactive environments \citep{yao2022react, shinn2023reflexion, huang2022language, ahn2022can, singh2022progprompt, reed2022generalist, driess2023palm, mu2023embodiedgpt, xiang2023language, zhangbuilding}.

However, existing benchmarks and training paradigms for such agents remain inadequate to study realistic test-time learning.
First, traditional benchmarks implicitly assume that learning does not happen at test time, serving instead as purely inference-time evaluations of static performance measures \citep{li2024embodied, cheng2025embodiedeval}. 
Second, even when learning is considered, it is not a continuous process during evaluation; rather, agents are trained over many episodes before evaluation \citep{shridhar2020alfworld, cote2018textworld, wang2025ragen, feng2025group}. Together, these paradigms obscure a central challenge faced by real-world agents: after initial training, learning must occur continuously throughout their lifetimes.

In this work, we study \textbf{test-time continual learning agents} that interact with the world, continuously acquire the necessary new knowledge and skills, and meaningfully improve during test-time deployment. This contrasts with traditional continual learning, which typically assumes a non-interactive setting and a clear boundary between training and testing \citep{mccloskey1989catastrophic, kirkpatrick2017overcoming, lopez2017gradient, shin2017continual}. Test-time continual learning can occur either without parametric updates, in which retrieval-based conditioning and in-context mechanisms drive learning, or with parametric updates, in which agents learn through test-time training.

Drawing inspiration from human cognitive development \citep{gopnik1997words,spelke2007core}, we hypothesize that successful test-time continual learning agents require five key abilities: \exploration\ \citep{berlyne1954theory,wilson2014humans}, \episodicmemory\ \citep{tulving2002episodic,schacter2007remembering}, \worldknowledge\ \citep{tse2007schemas,collins1975spreading}, \skilllearning\ \citep{anderson1982acquisition}, and \longhorizonplanning\ \citep{botvinick2009hierarchically,mattar2018prioritized}. They reinforce one another and jointly shape performance in long-horizon environments and tasks. Exploration and planning help agents gather novel experiences, from which they acquire world knowledge and skills that unlock future experiences, forming a positive feedback loop. Memory is central to this loop: past experiences, knowledge, and skills are often crucial for future success, making forgetting or poor generalization costly. Thus, memory cannot be a passive log of observations and actions; it must support learning by abstracting episodic experience into semantic knowledge and retaining causal, decision-relevant information for future planning.

To study the key abilities of test-time continual learning agents, we introduce \abbr{}, a novel evaluation framework that procedurally generates open-ended text games with rich, long-horizon environments. As shown in Figure~\ref{fig:teaser}, the generated games feature: (1) novel and diverse world knowledge about entities, such as areas, objects, and NPCs, as well as world dynamics that agents must explore and acquire; (2) rich game mechanics that encourage skill learning, such as note-taking; and (3) challenging long-horizon tasks that require agents to decompose objectives into subgoals, manage goals over time, and use episodic memory for more effective planning. Compared to video games \citep{chen2024can, hu2025lmgame}, computer tasks \citep{tan2024cradle, xie2024osworld, wang2025opencua}, and embodied 3D environments \citep{ren2025simworld, zhuang2025simworld, zhou2025virtual}, these text games isolate key agent abilities from other challenges, such as perception, visual grounding, and low-level control. They also allow agents to be studied in a highly controllable and reproducible setting while still capturing important dynamics of real-world decision-making.

In \abbr{}, we develop a game generation engine driven by an LLM-based entity and rule synthesis grounded in an ontology for long-horizon games as illustrated in Figure~\ref{fig:ontology_and_gen}. This engine can generate unlimited and diverse game entities, including locations, objects, and non-player characters (NPCs), as well as game rules, including action rules that define action effects and step rules that define environmental dynamics (e.g., NPC behavior). Critically, it also checks the game's correctness and repairs detected issues via program synthesis. Beyond game generation, we design multifaceted metrics to benchmark agent performance. Unlike typical game-based evaluation, we introduce a suite of diagnostic tests to probe an agent's abilities that are not directly reflected in game progress, such as world knowledge, episodic memory, object and action exploration, action diversity, and model cost, thereby further quantifying the agent's evolution over a long horizon.

In sum, our main contributions include: (1) \abbr{}, the first game generation engine for studying five key abilities of test-time continual learning agents; (2) a new multifaceted evaluation methodology for measuring game progress and diagnosing key abilities; (3) an empirical study of diverse agent paradigms, revealing critical limits in current agents, factors shaping their meaningful horizon, and a focused study of agent test-time training, centered on supervised fine-tuning (SFT) as a canonical instance of parametric weight updates, showing that despite the benefit of short-term memory, test-time parametric updates suffer from catastrophic forgetting, underscoring the importance of continual learning when updating weights.

%% file: sections/related.tex
\section{Related works}

\textbf{Text Games for Evaluating Agents.} There have been different kinds of environments for agent evaluation, including video games \citep{bellemare2013arcade, kempka2016vizdoom}, embodied simulators \citep{ren2025simworld, zhuang2025simworld, zhou2025virtual, li2024embodied, cheng2025embodiedeval}, web and OS environments \citep{yao2022webshop, zhou2023webarena, xie2024osworld}.
However, they do not fully support the evaluation of the five key abilities we propose for test-time continual learning agents. 
Specifically, it is difficult to generate novel world knowledge and skills that agents must learn, beyond the commonsense world knowledge and typical real-world skills that an LLM may already have learned. 
Also, success in these environments requires additional abilities such as perception, visual grounding, and mid-level or even low-level control, which are beyond the scope of our proposed test-time continual learning evaluation. 
Therefore, in this work, we focus on text games \citep{narasimhan2015language, he2016deep, cote2018textworld, shridhar2020alfworld, wang2023voyager, hu2025lmgame}. 
As shown in Table~\ref{tab:text_games}, \abbr{} combines scalable generation, action-independent dynamics, and long horizons. NetHack \citep{kuttler2020nethack} varies its dungeon while retaining a fixed, public set of entities and mechanics. TALES/Jericho \citep{cui2025tales,hausknecht2020interactive} includes fixed human-written games that may appear in the training data of current or future models. \abbr{} instead generates new entities, recipes, action rules, step rules, and quest dependencies, and uses world knowledge question-answering before gameplay to filter familiar games. Its diagnostic tools identify why an agent fails beyond game progress and reward. ByteSized32 \citep{wang2023bytesized32} also uses an LLM to generate text games, while \abbr{} generates games with long-horizon tasks.

\textbf{Test-Time Continual Learning Agents.}
Memory is central for agents to support test-time continual learning. It can be roughly categorized into 5 classes: \textbf{(1) RAG-based agents}: agents use retrieval-augmented generation (RAG) to store experience and knowledge in an external database as memory \citep{zhao2024expel, chhikara2025mem0, wang2023voyager}. \textbf{(2) Long Context agents}: agents append experience and knowledge to the context window at each step. 
\textbf{(3) Fixed Size Memory agents}: maintains a constant context length for memory. The simplest implementation is a sliding window of a fixed number of steps, serving as a short-term memory of past experience. The recent work MEM1 updates the memory using reasoning \citep{zhou2025mem1}, while MemAgent updates the memory using reinforcement learning (RL) \citep{yu2025memagent}.
\textbf{(4) Latent Memory agents}: implicit representations to encode and retrieve memory \citep{wang2024memoryllm, wang2025m+, zhang2025memgen}. 
\textbf{(5) Parametric Memory agents}: update the model's parameters via supervised fine-tuning (SFT) \citep{chen2023fireact} or RL \citep{wang2025ragen, feng2025group}. Other test-time training methods that model context dependencies by adapting part of the model’s weights at inference time are also highly relevant to our work \citep{sun2024learning, behrouz2025titans, zhang2025test, behrouz2025nested, tandon2025end}.

%% file: sections/env.tex
\section{\abbr{}}
\begin{figure*}[t]
    \centering
    \includegraphics[width=1\textwidth]{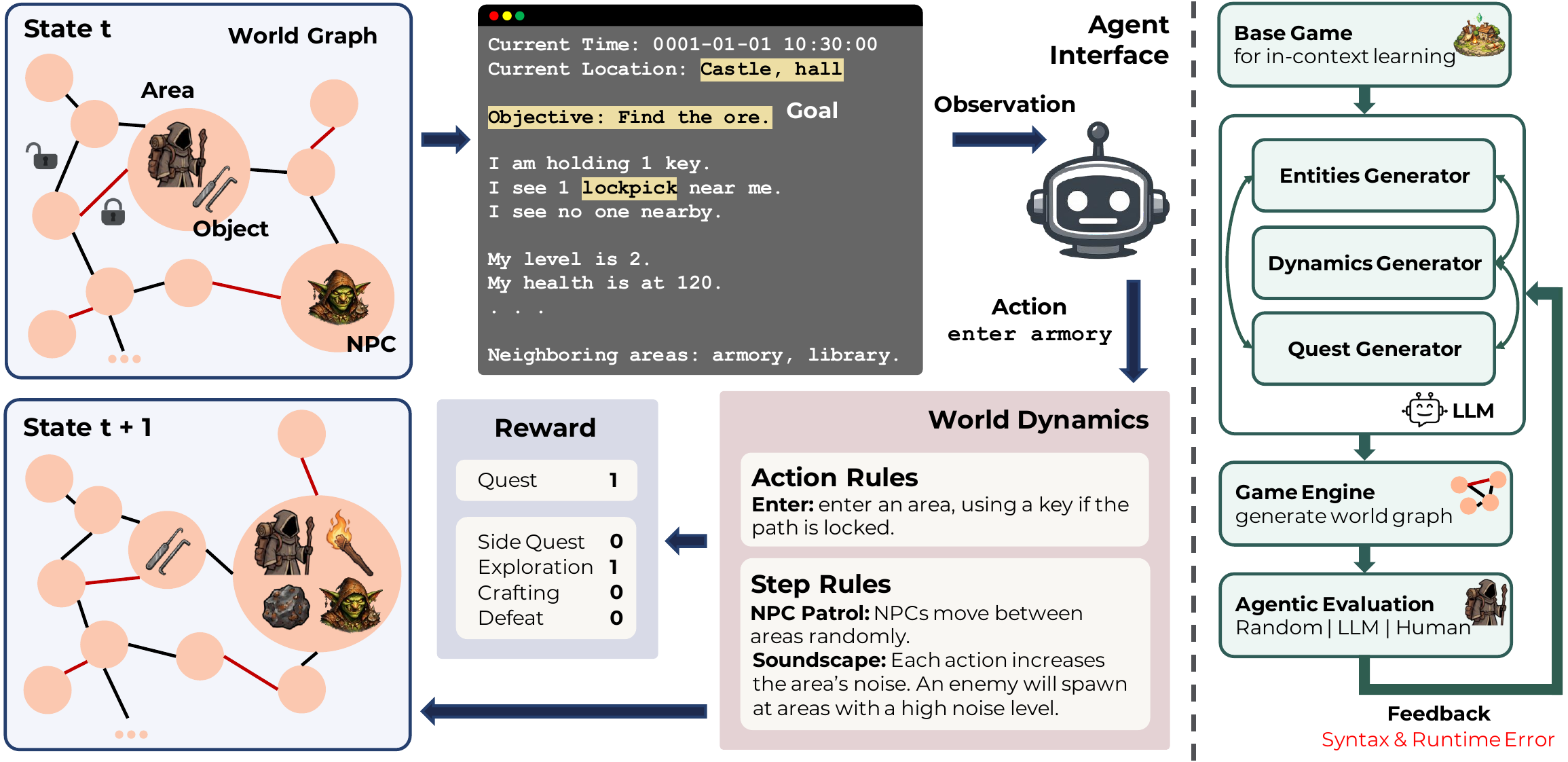}
    \caption{\textbf{\abbr{} game engine and generation pipeline.} At each step $t$, an agent observes a partial view of the world state as text, and makes a decision (i.e., an action). Action and step rules jointly update the world state from $t$ to $t+1$, and the agent receives rewards based on main quest progress and supplementary metrics, including side quest progress, exploration, crafting, and defeat. The left shows the game engine, and the right shows the entity, dynamics, and quest generators, each of which can access and modify other game elements.}
    \label{fig:ontology_and_gen}
\end{figure*}

In this work, we develop a novel agent evaluation framework, \abbr{}, to procedurally generate open-world text games for studying test-time continual learning agents, focusing on five key abilities. It is difficult to generate open-ended, long-horizon games in which success at later tasks depends on experience and learning from the past. Recent work has demonstrated that LLMs can generate text games end-to-end from scratch \citep[e.g.,][]{zhou2025story2game}. However, it faces several key challenges: (1) the game design to exercise 5 key abilities including how later tasks depend on previous exploration, experience, knowledge, and skills; (2) the control of important game elements (such as the number of entities and tasks) critical for diagnostic evaluation; (3) generating a large set of entities and world rules that the agent can meaningfully interact with or experience; (4) verification of the soundness of the game system; and (5) a consistent interface to AI agents, including truthful game state, feedbacks, and progression updates given agent actions.

Therefore, we propose a principled ontology for long-horizon game generation that formally specifies the fundamental components of the environment, ranging from entities to world dynamics. Figure~\ref{fig:ontology_and_gen} shows how these components support the engine and generation pipeline. At each step, the environment evolves through the composition of agent-initiated natural language actions (e.g., enter, store, attack) and environment-driven step rules (e.g., stochastic attacks on the agent by NPCs). Following each update, an agent receives a partial textual observation reflecting the current state of the environment and itself, as well as feedback. We introduce the formulation, game ontology, and the evaluation metrics in the sections below. More implementation details of \abbr{} are provided in Appendix~\ref{app:design_and_implementation}.

\subsection{Formulation}

\textbf{POMDP formalization.} 
We model the environment as a partially observable Markov decision process (POMDP) $(\mathcal{S}, \mathcal{A}, T, R, \Omega, O)$ with:
\begin{itemize}\setlength\itemsep{0pt}
  \item \textbf{State} $s_t \in \mathcal{S}$: structured world state containing a graph of locations, the distribution of objects, and NPC instances, as well as time and per-agent status (location, health, etc.). The agent additionally maintains an internal belief state about the world, updated based on partial observations.
  \item \textbf{Actions} $a_t \in \mathcal{A}(s_t)$: parameterized textual commands from a verb set with arguments.
  \item \textbf{Dynamics} $T(s_{t+1}\mid s_t,a_t)$: deterministic or stochastic updates induced by actions and step rules.
  \item \textbf{Observations} $O(s_t, a_{t-1}) \to o_t \in \Omega$: deterministic mapping that produces a natural language rendering of the current local state and feedback from the environment.
  \item \textbf{Reward} $R(s_t,a_t,s_{t+1})$: a signal reflecting high-level progress such as completing quest milestones, exploring new areas, crafting new objects, and defeating new enemies.
\end{itemize}
Time advances in fixed increments of $\Delta=10$ minutes of simulated time per step, yielding an explicit clock in the observation and enabling time-dependent rules.

\subsection{Ontology}
\label{sec:ontology}

\textbf{Overview.} We ground the game generation in the following ontology as illustrated in Figure~\ref{fig:ontology_and_gen}, which instantiates the key elements of a POMDP. We first sample three types of \textbf{game entities}: locations (i.e., places and areas), objects, and NPCs (non-playable characters). Their spatial relations are described in a \textbf{world graph}. The world graph at each step $t$ then becomes the state at that step. The \textbf{observation} for an agent is defined as its internal status and the visible part of the world graph. The \textbf{world dynamics} is jointly defined by a set of action rules and step rules. Each action rule defines the effect of an action that the agent can execute. Each step rule defines environmental changes (such as NPC behaviors and food respawns) that are triggered when world states meet specific conditions. Given a randomly sampled initial world graph, the game engine will simulate how the graph evolves to update the state according to world dynamics. We define a set of \textbf{goals} in the environment, which form main and side quests. We use several \textbf{rewards} measuring different aspects of the agent's game progress. We describe the key components below.

\textbf{Game Entities and World Graph.} The game world is instantiated from a declarative, easily modifiable specification that defines areas (e.g., castle hall), objects (e.g., wooden log), and NPCs (e.g., goblin). Each entity type is described by a set of attributes. In addition to names and levels, NPCs include properties such as health and attack strength, while objects specify craft ingredients, physical size, the areas in which they may be distributed, and more. Examples of each entity type are presented as a JSON object in Appendix \ref{app:example_game_entities}. Together, these elements form world graphs that represent the game's state at each step. Each node represents an area instance, and the object and NPC instances within it, and each edge denotes the connectivity between these areas, which may be locked or unlocked.

To define the game's initial state, we need to sample the initial world graph. The sampling process is conditioned on the entity level: higher-level NPCs (with greater strength, health, and richer inventories) and higher-level objects are more likely to appear in higher-level areas, thereby inducing a structured progression in environment difficulty and resource availability.

\textbf{Observations.}
The agent receives the observation at each step of the environment, consisting of: current time, current location, feedback, current status of the agent, including the objects in hand, equipment, level, attack, defense, health, and experience, as well as objects in the current area, NPCs in the current area, and neighboring locations. Refer to Figure~\ref{fig:ontology_and_gen} and Appendix~\ref{app:example_obs} for example observations.

\begin{figure*}[t]
    \centering
    \includegraphics[width=\textwidth]{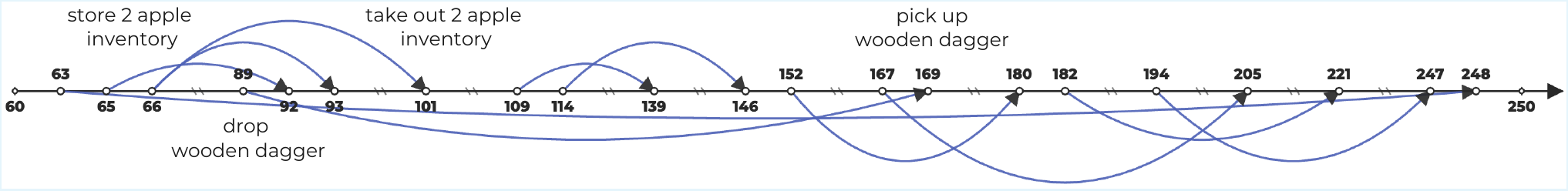}
    \caption{\textbf{Dependency graph.} \abbr{} features long-range action dependencies. For example, the agent picked up the wooden dagger at step 169 that was dropped at step 89.}
    \label{fig:dep_graph}
\end{figure*}

\textbf{World Dynamics.} Given the initial state defined by the initial world graph, the state and the underlying world graph are updated based on the world dynamics for the game engine. In particular, the world dynamics are implemented through a modular, two-stage rule system:
\begin{itemize}\setlength\itemsep{0pt}
  \item \textbf{Action rules} capture instantaneous, player-invoked operations (e.g., pick up or store objects). These step-level actions form long-range inter-dependencies that necessitate memory and learning when composed over time (see Figure \ref{fig:ontology_and_gen}). For example, objects dropped by an NPC defeated through combat may later become essential for crafting a weapon at a much later stage of the game. Successful game-play, therefore, requires long-horizon episodic memory over extended sequences of actions. We also visualize the dependency graph for analyzing long-range action dependencies over time (see Figure \ref{fig:dep_graph}). Refer to Appendix~\ref{app:example_action_rule} for the example of an action rule.

  \item \textbf{Step rules} encode persistent, stateful processes that are evaluated continuously as the environment evolves. Our environment includes numerous step rules that test an agent's memory and learning through indirect, underspecified environmental cues. For instance, a day-night cycle makes enemies stronger and more aggressive from 12:00 AM to 1:00 AM and weaker from 12:00 PM to 1:00 PM, thereby encouraging time-based strategic planning (e.g., staying at a safe location during that hour of the night). These temporal patterns are not explicitly disclosed to the player; they must be inferred from episodic experiences, such as changes in attack frequency or damage inflicted by nearby NPCs. Exploiting such rules thus requires accumulating and organizing semantic memory over a long horizon and inferring latent regularities from these episodic memories. Refer to Appendix~\ref{app:example_step_rule} for the example of a step rule.
\end{itemize}

In both cases, rules first check a set of preconditions over the current world and agent state, then apply state transitions that update entities, agent state, game progress, and finally emit feedback to the observation. While existing environments like TextWorld \citep{cote2018textworld} rely solely on deterministic rules, our game supports both deterministic and stochastic state transitions. For example, certain action rules (e.g., lock-picking) succeed with a defined probability, and step rules may introduce stochastic events such as spawning an NPC near the agent at midnight with a 50\% chance. This challenges simple memorization of deterministic dynamics, requiring agents to reason from episodic experience and update their internal state under uncertainty.

\textbf{Goals.} In our game, goals are formulated as quests. Each quest provides textual cues or instructions to guide the agent’s behavior, and delivers feedback and rewards upon completion. We consider two types of quests: main quests and side quests, both implemented via step rules. Main quests exhibit linear temporal dependencies, meaning that a goal can only be achieved after the preceding goal has been completed, thereby forming a coherent main storyline. In contrast, side quests have no preconditions and can be pursued at any time. Additional details about the quest design are provided in Appendix~\ref{app:tasks}.

\textbf{Rewards.} The game provides multiple reward signals. We define the \texttt{quest} reward as the number of completed main quest stages. In addition, we introduce several supplementary rewards: the \texttt{side quest} reward, defined as the number of completed side quests; \texttt{exploration}, defined as the number of explored areas; \texttt{craft}, defined as the number of unique objects crafted; and \texttt{defeat}, defined as the number of unique NPCs defeated.

\subsection{Game generation with program synthesis}
\label{sec:game_generation}
We develop game generators through LLM-based program synthesis and editing. The framework is based on Aider \citep{aider2026} and consists of three components: an entity generator, a rule generator, and a quest generator. Each generator is conditioned on an example game (i.e., base game) and produces new entities, dynamics, and quests that modify and extend it. Human design is involved only in constructing the base game, which is written to test the agent's five key abilities while maintaining a clear structure that enables the LLM to learn from context and generate new games. Generated games contain substantially different entities, storylines, actions, and world dynamics from the base game. \abbr{} relies only on a minimal RPG-style scaffold, namely abstract classes such as agents, areas, objects, and NPCs, rather than hand-designed game content or fixed RPG conventions. The concrete attributes of entities and the mechanics governing interactions are generated by LLMs, including action effects, entity dependencies, NPC behaviors, stochastic outcomes, and long-range consequences, resulting in games with distinct gameplay loops that must be learned through interaction rather than solved by relying solely on prior knowledge of RPGs. The base game also incorporates additional features, including a tutorial room, physical world alignment, and online expansion, which are inherited by subsequently generated games. Further details on these additional features are provided in Appendix~\ref{app:additional_features}.

However, the generated games are not guaranteed to be sound. We assess quality through static checks, execution tests, and human playtesting. An automated feedback loop runs each game with arbitrary agents (e.g., agents that take random actions) and reports errors for repair. This enables run-time functional validation beyond static syntax checking. The pipeline is shown in Figure~\ref{fig:ontology_and_gen}, and example programs are provided in Appendix~\ref{app:example_program_synthesized}. Appendix~\ref{app:generator_evaluation} reports experiments with open-weight and proprietary generators and a qualitative analysis of generation failures.

\begin{figure*}[t]
    \centering
    \includegraphics[width=\textwidth]{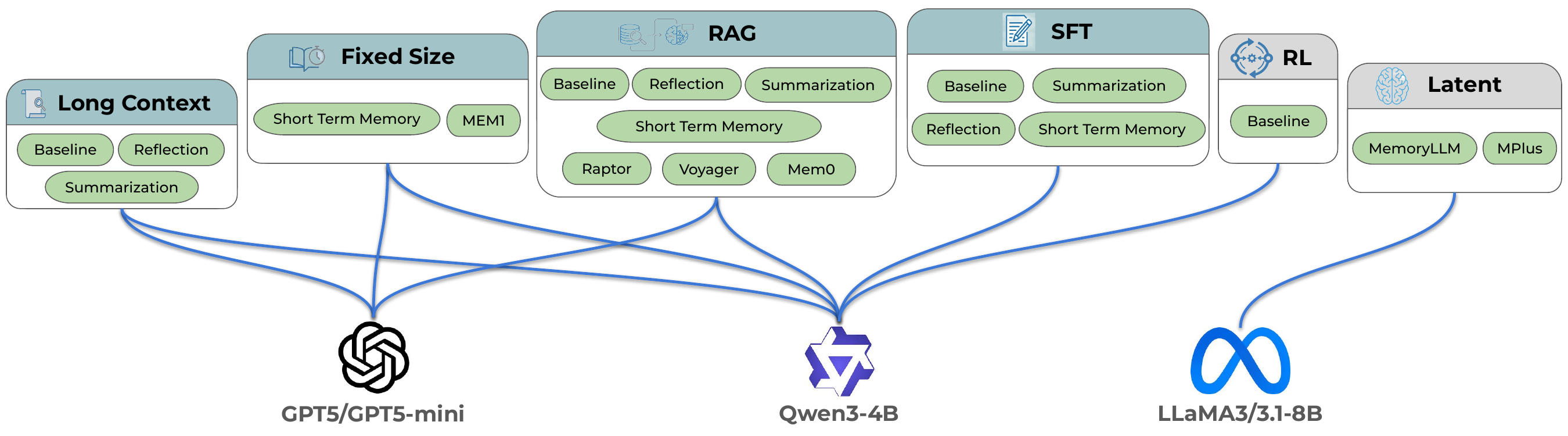}
    \caption{\textbf{Agent taxonomy.} 6 method paradigms and 3 language model backbones. We focus the evaluation on Long Context Agents, Fixed Size Memory Agents, RAG Agents, and SFT Agents, optionally augmenting them with additional mechanisms such as reflection, summarization, and short-term memory.}
    \label{fig:agent_taxonomy}
\end{figure*}

\subsection{Evaluation metrics}
\label{sec:evaluation_metrics}
We evaluate agents with three types of metrics:

\textbf{Game Progress.} We use the rewards defined in Section~\ref{sec:ontology} to measure game progress. The main reward is the \texttt{quest} reward, and the supplementary reward is the total of \texttt{side quest}, \texttt{exploration}, \texttt{craft}, \texttt{defeat} rewards. To visualize the main and supplementary game progress rewards together and compare different methods, we normalize both rewards to account for differences in scale across runs within the same game. Let $R^{\text{main}}_t$ and $R^{\text{sup}}_t$ denote the cumulative rewards up to step $t$ for the main and supplementary components, respectively. We compute reference maxima across runs within that game, $M_{\text{main}} = \max_{\text{runs},\, t} R^{\text{main}}_t$ and $M_{\text{sup}} = \max_{\text{runs},\, t} R^{\text{sup}}_t$. The normalized combined reward is then defined as
$R^{\text{combined}}_t
=
\frac{1}{2}
\left(
\frac{R^{\text{main}}_t}{M_{\text{main}}}
+
\frac{R^{\text{sup}}_t}{M_{\text{sup}}}
\right).$
This normalization ensures comparable contributions from both components despite differing magnitudes, while keeping the combined metric bounded for evaluating agents with different LLMs. For game progress comparison, performance is compared primarily using the main \texttt{quest} reward. When multiple runs achieve the same main quest reward, the tie is broken by comparing the total supplementary reward.

\textbf{Diagnostic Testing.} We further propose a suite of diagnostic tests to evaluate key abilities beyond the reward signals that are directly reflected in game progress:
\begin{itemize}\setlength\itemsep{0pt}
    \item \textbf{World Knowledge QA:} Multiple choice questions are generated using rule-based templates for crafting recipes, object distributions, and spatial connectivity. LLMs are used only for world-dynamics questions and distractor enrichment, conditioned on the game definition and code. Examples are shown in Tables~\ref{tab:game_1_stats} and~\ref{tab:game_2_stats}. We evaluate accuracy before and after gameplay, and progressively during exploration, to measure knowledge acquired through interaction. To ensure novelty of the world knowledge in a game, we tested pretrained LLMs on World Knowledge QA before gameplay and filtered out any game on which accuracy was above a threshold. This means that the world knowledge in the game differs from the commonsense knowledge already acquired by an LLM.

    \item \textbf{Episodic Memory QA:} Multiple choice questions are constructed from the agent's trajectory in the game world using rule-based templates, including visited areas, crafted and acquired objects, dropped objects, defeated NPCs, and temporally ordered actions. These questions evaluate the agent’s episodic memory of its own past experiences. Examples are shown in Table~\ref{tab:game_1_stats} and Table~\ref{tab:game_2_stats}.

    \item \textbf{Object and Action Exploration:} Beyond area exploration reflected in game progress, objects and actions are central to successful gameplay, and effective agents must explore both extensively. We therefore report the proportion of objects acquired (defined as picked up or stored) and the proportion of available actions executed by the agent as proxy metrics for its exploration capability.

    \item \textbf{Action Diversity:} We quantify action diversity over the agent’s action history using entropy computed within a sliding window. An effective agent should maintain sufficiently diverse behavior, rather than repeatedly executing a small set of actions or exhibiting a sharp decline in diversity over time. We calculate the entropy of a window of actions, normalized to $[0, 1]$, as the action diversity score ($AD$). Here, $\mathrm{AD}=-\frac{\sum_{i=1}^{N} p_i \log p_i}{\log N}$, where $N$ denotes the total number of available actions and $p_i$ denotes the empirical probability of action $i$ in the window.
\end{itemize}

We manually checked 30 sampled questions from Remnant across both QA types and found them 100\% correct. Craft and defeat rewards measure successful skill use, while main and side quest progress measure planning and goal switching. Appendix~\ref{app:evaluation_validation} gives the full ability mapping and validation details.

\textbf{Model Cost.} The sum of input tokens and output tokens (i.e., total tokens) used for each agent.

%% file: sections/agent.tex
\section{Agent paradigms}

We implement a \textbf{universal agent interface} to evaluate a range of LLM-based agents with different base models, grouped into 6 paradigms, as well as two additional baselines: one with no memory and one that acts randomly.

\textbf{Long Context (LC) Agents} append observations, reasoning, and actions to their context at each step.

\textbf{Fixed Size Memory Agents} maintain a fixed-size context as memory: either through a sliding window as short-term memory (STM) or a bounded, self-updating memory buffer \citep{zhou2025mem1}.

\textbf{RAG Agents} store observation-reasoning-action tuples in an external database of embedding-text pairs. During inference, the agent retrieves several relevant entries and appends them to the model context for decision-making. We adapt four variants: Vanilla RAG~\citep{lewis2020retrieval}, Mem0~\citep{chhikara2025mem0}, Raptor~\citep{sarthi2024raptor}, and Voyager~\citep{wang2023voyager}.

\textbf{SFT Agents} encode experience (i.e., observation-reasoning-action tuple) into parameters via LoRA-based supervised fine-tuning~\citep{hu2022lora}, updating adaptor weights.

\textbf{RL Agents} update the LoRA adapter weights with PPO \citep{schulman2017proximal, feng2025group}, using environment rewards defined in Section~\ref{sec:ontology}.

\textbf{Latent Agents} compress experience into learnable latent memory tokens integrated into the model’s hidden states, enabling persistent storage and retrieval, as seen in MemoryLLM \citep{wang2024memoryllm} and MPlus \citep{wang2025m+}.

All agents adopt the ReAct prompting method~\citep{yao2022react}, and we use the same final-stage prompt to output an action from the action space, as shown in Appendix~\ref{app:example_agent_prompt}. We focus the evaluation on Long Context Agents, Fixed Size Memory Agents, RAG Agents, and SFT Agents and optionally augment them with additional mechanisms, including reflection (REFL), summarization (SUM), and short-term memory (STM) which is a fixed-size context window that stores a specified number of the most recent observation–reasoning–action tuples \citep{shinn2023reflexion, lee2024human}. More details are provided in Appendix~\ref{app:additional_agent_implementation}. We evaluate 2 representative classes of LLM backbones for agents, namely GPT-5 \citep{singh2025openai} and Qwen-3 \citep{yang2025qwen3}, covering both proprietary and open-weight models. As some methods learn purely from context, while others require parameter updates, we present the agent taxonomy in Figure~\ref{fig:agent_taxonomy} to clarify which LLM backbones are associated with each method. Note that we also include LLaMA-3 \citep{grattafiori2024llama} because MemoryLLM and MPlus are trained models and are only compatible with LLaMA 3/3.1-8B.

%% file: sections/exp1.tex
\section{Experiment 1 - diagnosing five key abilities of agents}
\label{sec:case_study}

\begin{figure*}[t]
    \centering
    \includegraphics[width=\textwidth]{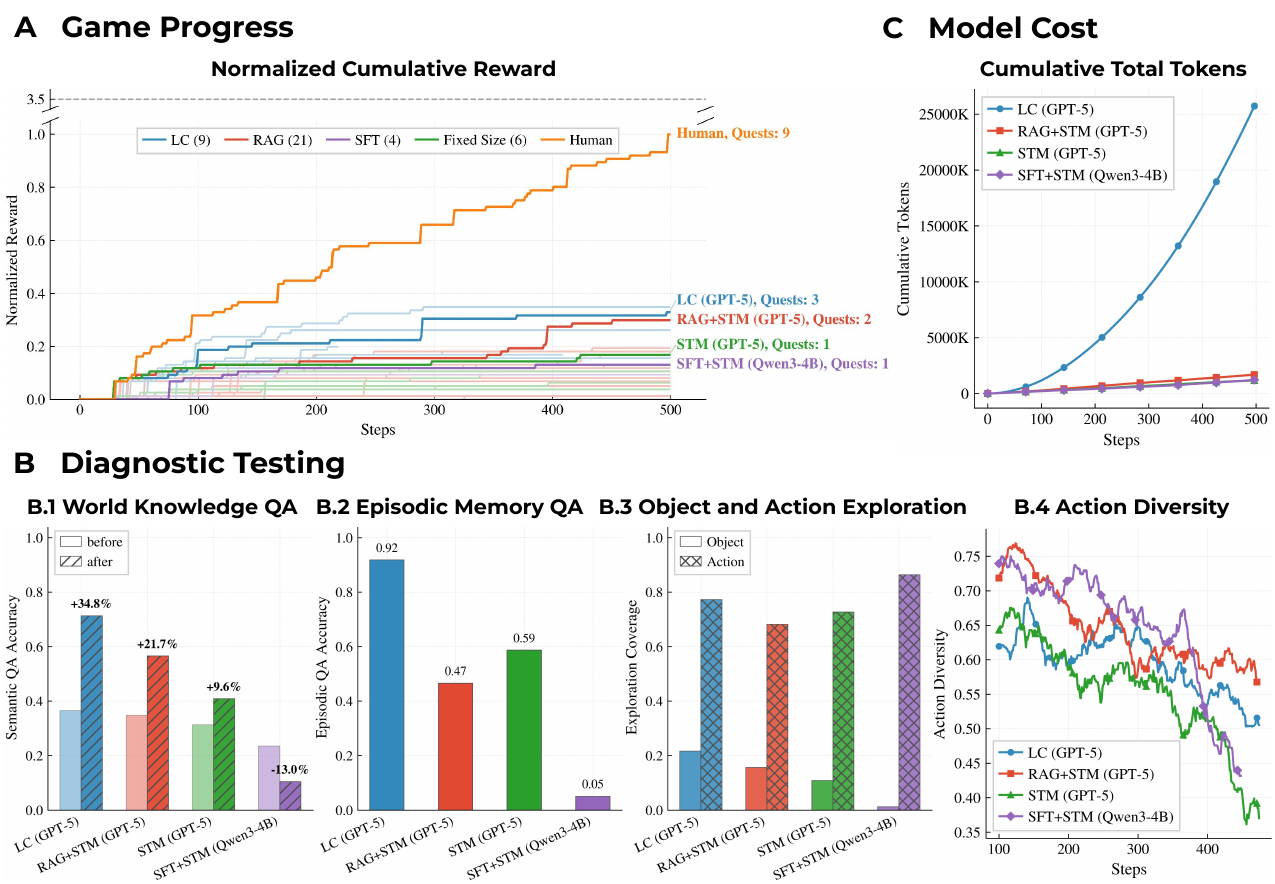}
    \caption{\textbf{Experimental results for Experiment 1 on Remnant.} In \textbf{A Game Progress}, the best-performing agent in each paradigm is highlighted, and the dashed line indicates the theoretical maximum reward. Performance is ranked by main quest reward, with supplementary reward breaking ties. Long Context GPT-5 performs best but remains below human performance. In \textbf{B Diagnostic Testing} and \textbf{C Model Cost}, we visualize the best-performing agents. The strong correlation with game progress is observed for agents without parameter updates at test time. The Long Context agent has a quadratic token cost.}
    \label{fig:vis_results_exp1}
\end{figure*}

To investigate the 5 key abilities of test-time continual learning agents, we employ \abbr{} to generate a long-horizon game environment with the characteristics illustrated in Figure~\ref{fig:teaser}.

\subsection{Game description}
Remnant contains 18 areas, 83 object types, and 13 NPC types. It also has 24 stages in the main quest. The complete specifications are provided in Table~\ref{tab:game_1_stats} in Appendix~\ref{app:generated_game_details}. We visualize a partial dependency graph of agent interactions in this environment to illustrate long-range action dependencies over time (see Figure~\ref{fig:dep_graph}). In addition, the environment includes a 26-step tutorial designed to familiarize agents with the action space. Further details of the tutorial stage are provided in Appendix~\ref{app:additional_features}.

\subsection{Analysis and discussion}
We run the agents for 500 steps and visualize the game progress via normalized cumulative reward in Figure~\ref{fig:vis_results_exp1}. We first find that the Long Context agent has the highest performance (with GPT-5), showing that \textbf{(1) the performance of the agent scales with the amount of information stored in the memory and during inference} because RAG-based methods store everything in the external database, but only retrieve a fixed amount of information to be conditioned during inference. Also, short-term memory stores only a fixed amount of information, which influences decision-making. Additionally, \textbf{(2) the performance of the Long Context agent scales with the LLM's long-context modeling and reasoning abilities} because the performance degrades when pairing the agent with GPT-5-mini, and even worse with Qwen3-4B. We therefore conduct experiments with additional frontier LLMs using the Long Context agent. In Appendix~\ref{app:more_results}, Figure~\ref{fig:cumulative_reward_lc_all_llms} shows that the best-performing LLM is Claude-Opus-4.6, but it still lags behind human performance. A critical limitation for Long Context agents is the context window: as the number of input tokens grows linearly with the number of steps, the agent stops running once reflection is added because it incurs extra tokens per step. Therefore, the context window limits the meaningful horizon for these agents.

 To diagnose the best-performing agent in each category, we also visualize the diagnostic metrics in Figure~\ref{fig:vis_results_exp1} and observe a strong correlation to game performance for agents without parameter updates at test time. SFT is the exception, as discussed in Section~\ref{sec:agent_mechanisms}. For instance, Long Context GPT-5 improves its World Knowledge QA accuracy from 0.365 to 0.713 after gameplay and achieves the highest Episodic Memory QA accuracy among the highlighted agents. This demonstrates that it acquires semantic world knowledge and retains episodic experience more effectively than other agents. The Long Context agent stores all past experience as text in its context, so World Knowledge QA and Episodic Memory QA reduce to long-context reasoning. To further understand how world knowledge is acquired during gameplay, we conduct World Knowledge QA every 100 steps from the start of the game for the Long Context agent with GPT-5. Figure~\ref{fig:progressive_wk_qa} in Appendix~\ref{app:more_results} shows that world knowledge increases from step 0 to step 300, then gradually flattens out due to a lack of further exploration. These results provide direct evidence of test-time learning and interleaved learning and inference. Moreover, the Long Context agent exhibits the strongest object exploration ability among the 4 agents, primarily because its memory records all past explorations, enabling it to identify unexplored objects. However, all agents have room to improve in exploring objects and actions. The action diversity plot also shows that the Long Context agent, although its diversity decreases over time, still maintains a wide range of actions throughout the long trajectory and does not collapse to a single action or a small subset of actions. On the other hand, the STM agent and the SFT agent exhibit a sharp decrease in action diversity, which coincides with the plateau observed in the cumulative reward. This suggests a correlation between action diversity and game performance, indicating another barrier to achieving a meaningful horizon for agents.
 
 Despite the strong performance of the Long Context agent, its model cost, measured by cumulative token count, increases quadratically with the number of steps. Hence, its meaningful horizon will be reduced under a budget limit, whereas other types of agents use many fewer tokens (i.e., linearly with respect to the number of steps), leading to a longer horizon under a fixed budget.

Tables~\ref{tab:results_proprietary_exp1} and~\ref{tab:results_open_exp1} report full Experiment 1 results, and Table~\ref{tab:multiple_runs_variance} reports three repeated runs, all in Appendix~\ref{app:more_results}. Using these results and trajectory visualizations from Appendix~\ref{app:additional_features}, we identify five failure patterns across agent paradigms and models, corresponding to the five key abilities studied in this work, followed by an analysis of model cost and reasoning efficiency:
\begin{itemize}\setlength\itemsep{0pt}

    \item \textbf{\Exploration: Insufficient Object and Action Coverage.}
    Agents exhibit limited exploration over both objects and the available action space. In particular, they often decline to collect objects that are not directly related to the current quest objective, even when such objects serve as intermediate crafting ingredients required for important objects. This myopic exploration strategy restricts the agent’s ability to acquire prerequisite resources and undermines long-term planning. Moreover, all evaluated agents fail to systematically explore the full action space, thereby preventing them from learning the causal effects and affordances of certain actions. The absence of such exploratory behavior constrains future decision-making, as potentially beneficial actions remain unexplored.

    \item \textbf{\EpisodicMemory: Repetitive Behaviors, Limited Failure Recovery, and Hallucination.}
    Agents frequently select actions that are either uninformative or explicitly invalid according to environment feedback. A common failure mode is the emergence of local-minimum behavioral loops, in which the agent repeatedly executes near-identical action sequences accompanied by similar reasoning traces, despite clear negative performance signals. This is also shown in the action diversity plot in Figure~\ref{fig:vis_results_exp1}, where agents exhibit decreasing action diversity over time. For instance, agents may persistently attempt to enter restricted areas during combat, even after receiving repeated error messages indicating that such actions are disallowed. Notably, this issue is observed even in Long Context agents, which encounter the same corrective feedback multiple times within their context window yet continue to repeat the erroneous behavior. We hypothesize that this limitation is associated with deficiencies in the use of episodic memory: an ideal agent should reflect on failure experiences, revise the plan, and adapt subsequent decisions accordingly. In addition, agents occasionally fail to recognize objects that are present in the current environment, instead initiating unnecessary search behaviors elsewhere. They may also lose track of object locations over multiple steps, despite having previously observed them. Such patterns suggest episodic hallucination, whereby the agent’s internal memory representation diverges from the actual interaction history.

    \item \textbf{\WorldKnowledge: Semantic Memory Hallucination.}
    Agents, particularly those powered by smaller language models, frequently exhibit hallucinations in semantic knowledge. For example, they may attempt to craft nonexistent objects, apply incorrect recipes, or fail to reason about, retain, and leverage the game environment's dynamics. Furthermore, even when provided with correct information in the current observation, they sometimes fail to update their internal world knowledge accordingly. These limitations significantly hinder the agent’s ability to learn and adapt continuously over time.

    \item \textbf{\SkillLearning: Inefficiency and Failure to Acquire Procedural Skills.}
    Even when an adversary's behavior follows a deterministic, repetitive pattern, most agents fail to acquire an effective procedural counter-strategy and instead rely on short-horizon reactive decisions. Only the Long Context agent exhibits partial adaptation by forming a rudimentary combat pattern, although the resulting policy remains suboptimal. Furthermore, none of the evaluated agents acquire auxiliary skills that could improve task efficiency, such as externalizing crafting recipes into written notes to facilitate future planning and execution. These observations suggest that continual learning agents should extend beyond merely accumulating declarative world knowledge to include the acquisition and consolidation of procedural skills, thereby forming a more robust procedural memory.
    
    \item \textbf{\LongHorizonPlanning: Poor Goal Maintenance and Switching.}
    Agents interact in environments with multiple concurrent objectives and internally generated subgoals derived from task decomposition. This results in several parallel goals during interaction. While agents can execute immediate subgoals effectively, they frequently fail to re-anchor on the primary objective after completing them. Moreover, there is little evidence of explicit reasoning that supports a switch from one goal to another. These observations suggest that continual learning agents should maintain persistent goal representations while enabling flexible prioritization and switching across both internal subgoals and external objectives, thereby supporting coherent long-horizon planning and execution.

    \item \textbf{High Model Cost and Inefficient Reasoning.}
    Finally, we observe that many agent paradigms and models rely on excessive context and reasoning tokens. This not only raises inference costs but also slows decision-making. Future work should focus on developing agents that (1) use a fixed-size context as working memory for direct conditioning, (2) consolidate new experiences and knowledge into the model weights, and (3) enable language models to reason more efficiently in agentic tasks by reducing reasoning tokens.

\end{itemize}

We further observe that short-term memory consistently improves both RAG and SFT agents. Although the game requires long-horizon planning, agents also need working memory to maintain short-term goals. For example, when the goal is to collect five wooden logs, the agent must remember this goal and track how many logs have been collected at each step. Moreover, Table~\ref{tab:stm_size_analysis} in Appendix~\ref{app:more_results} shows that both game-progress performance and World Knowledge QA accuracy increase monotonically with short-term memory size. We also find that reflection and summarization do not help reasoning-model agents. We therefore hypothesize that these models already implicitly reflect on their experiences and summarize key points during reasoning.

\subsection{Evaluation across multiple games}
\label{sec:multi_game}
We compare four representative agents across all six generated games, with up to 500 steps per run. The four other games had already been generated for the diversity analysis in Appendix~\ref{app:generated_game_details}. We conducted 16 additional runs on these four games plus three GPT-5 runs on Mark. The SFT+STM result on Mark is from Table~\ref{tab:results_exp2}. Table~\ref{tab:multi_game_results} reports completed/total main quest stages and supplementary reward. Mean quest completion is the arithmetic mean of the six completion fractions.

\input{sections/multi_game_table}

Long Context ranks first in all six games and completes Mark in 192 steps. RAG+STM ranks second in five. Supplementary reward resolves its ties with Fixed Size STM in Mark and Quarantine. Fixed Size STM and SFT+STM tie for second in Robot Kingdom. The overall order matches Experiment 1, with mean quest completion of 47.7\%, 30.0\%, 24.1\%, and 19.3\%, respectively. Across the five games other than Remnant, Long Context averages 51.6k tokens per step versus 2.5--3.4k for the others, averaging the per-game token means. Full results are in Appendix~\ref{app:multi_game_results}.

Three repeated runs of the GPT-5 RAG baseline on Remnant each complete one main quest stage, with supplementary rewards of 4, 5, and 4 (Table~\ref{tab:multiple_runs_variance} in Appendix~\ref{app:more_results}). These runs assess variability on the same game, while the multi-game results assess variability across different entities, rules, and quests.

%% file: sections/multi_game_table.tex
\begin{table}[t]
\centering
\small
\setlength{\tabcolsep}{3pt}
\caption{\textbf{Agent evaluation across six generated games.} Entries give completed/total main quest stages, with supplementary reward $SQ+E+C+D$ in parentheses. Performance is ranked by main quest reward, with supplementary reward breaking ties. Mean quest completion averages the six completion fractions. The highest main and supplementary rewards in each game and the highest mean are bold. All runs have a budget of up to 500 steps.}
\label{tab:multi_game_results}
\resizebox{\textwidth}{!}{%
\begin{tabular}{lrrrrrrr}
\toprule
\multirow{2}{*}{Representative agent} & \multicolumn{6}{c}{Game} & \multirow{2}{*}[-3pt]{\shortstack{Mean quest\\completion}} \\
\cmidrule(lr){2-7}
& Remnant & Mark & Metropolis & Saltglass & Quarantine & Robot Kingdom & \\
\midrule
Long Context (GPT-5) & \textbf{3}/24 (13) & \textbf{17}/17 (20) & \textbf{9}/21 (14) & \textbf{6}/8 (\textbf{26}) & \textbf{13}/28 (\textbf{22}) & \textbf{2}/21 (\textbf{13}) & \textbf{47.7\%} \\
RAG+STM (GPT-5) & 2/24 (\textbf{15}) & 11/17 (\textbf{31}) & 8/21 (\textbf{17}) & 4/8 (10) & 4/28 (13) & 1/21 (5) & 30.0\% \\
Fixed Size STM (GPT-5) & 1/24 (9) & 11/17 (22) & 4/21 (9) & 3/8 (12) & 4/28 (4) & 1/21 (8) & 24.1\% \\
SFT+STM (Qwen3-4B) & 1/24 (6) & 7/17 (16) & 7/21 (6) & 2/8 (8) & 2/28 (8) & 1/21 (8) & 19.3\% \\
\bottomrule
\end{tabular}%
}
\end{table}

%% file: sections/exp2.tex
\section{Experiment 2 - focused study on agent test-time training}
\label{sec:agent_mechanisms}

Unlike non-parametric memory paradigms, such as Long Context, Fixed Size, and RAG, which store experience in context or external memory, parametric test-time training internalizes experience into the model’s own weights. Although the SFT agent evaluated in Experiment 1 is not the best-performing approach, it warrants focused study as the most direct instance of parametric test-time training, especially given the growing interest in test-time training \citep{sun2024learning, zhang2025test, tandon2025end}. However, Qwen3-4B is insufficiently capable for the game used in experiment 1, so we generate a simpler game to verify our findings and hypotheses on agent test-time training. Specifically, this game features simpler main quests, reduced crafting hierarchies, and weaker enemies. We also disable the side quests in the game to avoid goal switching between the main and side quests for agents.

\begin{figure*}[t]
    \centering
    \includegraphics[width=\textwidth]{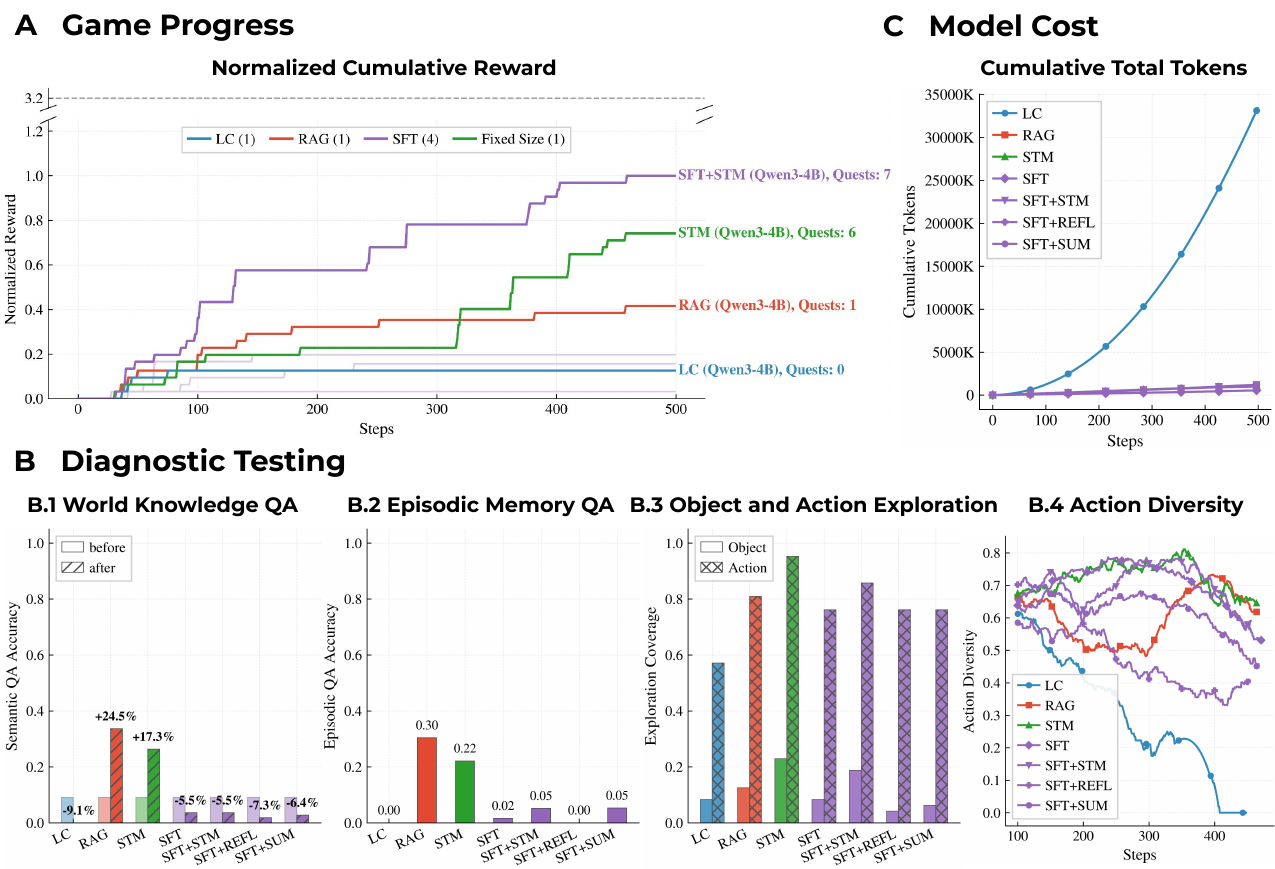}
    \caption{\textbf{Experimental results for Experiment 2 on Mark with Qwen3-4B.} In \textbf{A, Game Progress}, the best-performing agent in each paradigm is highlighted, and the dashed line indicates the theoretical maximum reward. SFT with short-term memory performs best. In \textbf{B, Diagnostic Testing} and \textbf{C, Model Cost}, we visualize all evaluated agents. SFT is an exception to the correlation observed for agents without parameter updates. Its QA degradation is not reflected in quest progress.}
    \label{fig:vis_results_exp2}
\end{figure*}

\subsection{Game description}
Mark contains 14 areas, 49 object types, and 12 NPC types. It also has 17 stages in the main quest. The complete specifications are provided in Table~\ref{tab:game_2_stats} in Appendix~\ref{app:generated_game_details}. In addition, the environment includes a 28-step tutorial designed to familiarize agents with the action space. Further details of the tutorial stage are provided in Appendix~\ref{app:additional_features}.

\subsection{Analysis and discussion}
We evaluate four SFT agents for 500 steps, including a vanilla baseline and three variants with augmented mechanisms. In addition, we include comparisons with other baselines: a Long Context agent, a RAG agent, and a Short-Term Memory (STM) agent. All agents are powered by Qwen3-4B to ensure fair comparisons. Same as experiment 1, we visualize the game progress, diagnostic testing, and model cost in Figure~\ref{fig:vis_results_exp2}. In Appendix \ref{app:more_results}, Table~\ref{tab:results_exp2} shows the full results in Experiment 2.

The experimental results indicate that incorporating short-term memory substantially improves the SFT agent's performance, yielding the best overall results in the group. Moreover, the SFT agent augmented with short-term memory outperforms the vanilla Short-Term Memory agent. This suggests that short-term memory provides a complementary working memory for tracking short-term goals, while \textbf{test-time parametric weight updates act as a form of long-term memory that consolidates experience}. The matched-game comparison in Table~\ref{tab:multi_game_results} supports the role of backbone capability. GPT-5 Long Context completes all 17 stages of Mark in 192 steps, whereas Qwen3-4B Long Context completes 0/17 in 500 steps. In addition, reflection and summarization do not improve the SFT agent. These findings are consistent with experiment 1.

In the diagnostic evaluation, the Long Context agent shows zero accuracy on both the World Knowledge QA and Episodic Memory QA after gameplay, and its action diversity declines progressively, ultimately converging on a single action. These results suggest that the Long Context agent undergoes collapse and degeneration. Furthermore, in line with the observations from experiment 1, SFT agents show decreased World Knowledge QA accuracy after training and very low Episodic Memory QA accuracy. We hypothesize that this degradation arises from reduced general language capability due to training, i.e., catastrophic forgetting. Thus, although short-term memory benefits agent test-time training, the parametric updates that serve as its long-term memory may suffer from catastrophic forgetting, highlighting the importance of \textbf{continual learning when updating weights at test time}. Crucially, effective test-time training must not only mitigate forgetting but also preserve the model's plasticity, enabling it to continue acquiring new knowledge and skills, as a learner that forgets nothing yet learns nothing is of little use. Because our diagnostics quantify world knowledge and episodic memory independently of the learning algorithm, they apply to any test-time training method, and a comprehensive comparison across other parametric methods is a natural extension that \abbr{} enables. Future work on agent test-time training should therefore develop continual learning algorithms that balance this \textbf{stability--plasticity trade-off}.

%% file: sections/appendix.tex
\clearpage
\beginsupplement

\begin{center}
     \Large\textbf{Appendix}
\end{center}

\input{sections/results_table}
\input{sections/additional_results}

\section{Environment design and implementation details}
\label{app:design_and_implementation}

We provide more details below on the environment design and implementation discussed in Section~\ref{sec:ontology}.

\input{sections/evaluation_validation}

\subsection{Tasks}
\label{app:tasks}

\paragraph{Main quest.}
Following standard game design, the main quest is structured as a sequence of interdependent chapters, each comprising multiple stages (i.e., tasks) that evaluate the agent's five key abilities. Progression is strictly sequential: a new stage becomes available only after the preceding one has been completed. A representative example is defeating a boss, which typically requires several intermediate sub-tasks, such as exploring new areas to gather objects, crafting appropriate weapons and armor, and acquiring combat skills. Completing these tasks requires the agent to continuously explore and acquire world knowledge and skills while retaining episodic experiences and maintaining coherent long-horizon objectives.

\paragraph{Side quest.}
Side quests provide auxiliary objectives with shorter horizons than the main quest and can be completed in any order. Each area unlocks four types of side quests, namely collect, talk, craft, and trade, which draw on the world knowledge and experiences accumulated through prior interactions. For example, finding the target NPC to talk to involves remembering where that NPC was previously encountered. Completing the task yields rewards that facilitate progress in the main quest. At the same time, side quests require the agent to temporarily shift the goal away from its main objective toward other tasks, thereby extending the effective decision horizon and evaluating the agent’s ability to maintain and switch goals.

\subsection{Game generation}
\label{app:game_generation}

\paragraph{Entity generation.}
A world definition JSON file specifies entities and their attributes, such as the economic value of objects and the attack power of NPCs. To expand the state space with new entities, we employ an entity generator that performs in-context learning over the existing world definition. During generation, we impose hard semantic constraints and balancing heuristics to preserve coherence and a well-shaped state distribution. For example, a coral reef would not be assigned to the library, and object levels are distributed approximately uniformly across the available range. To guide the process, the generator receives not only the current world definition but also an analysis chart summarizing entity distributions, such as the number of objects by type and the counts of enemy versus friendly NPCs. The generator can additionally access and modify world dynamics, world graph instantiation, and quest objectives, ensuring that newly introduced entities remain consistent with the existing game. By default, the entity generator is powered by a coding agent; however, we also provide a fallback to direct LLM-based generation, since the coding agent may occasionally fail to produce valid code edits to the JSON file.

\paragraph{Rule generation.}
We employ two sub-generators to synthesize action rules and step rules using in-context learning from the base game. Both rule generators can inspect and modify the world definition and the world graph instantiation process. This design allows the state space and the world dynamics to evolve together, as new entities are introduced alongside the rules governing object affordances and NPC behaviors. Therefore, our generated games feature rich dynamics that apply to a diverse set of entities.

\paragraph{Quest generation.}
We additionally introduce a dedicated quest generator that expands or modifies the main storyline by producing quest chapters composed of multiple stages. Unlike recent work that primarily focuses on generating coherent narratives \citep{zhou2025story2game}, our quest generation framework emphasizes objective diversity, hierarchical goal decomposition, long-horizon dependencies, and calibrated difficulty progression to evaluate the five key abilities of a test-time continual learning agent.

\paragraph{Game-specific vocabulary.}
The unfamiliar semantics are deliberate. Generated games should contain knowledge that the evaluated LLM cannot retrieve directly and must instead learn through interaction with the environment. Natural language descriptions and observed effects explain what unfamiliar objects do. \abbr{} allows the generator prompt to favor interpretable names and transparent compounds when desired. At the same time, as in many computer games, invented terms can be a meaningful part of the environment. The agent must acquire game-specific vocabulary from interactions during gameplay, for example by inspecting \texttt{gavelwood\_spire}. World Knowledge QA and interaction traces help determine whether the agent acquired those semantics before a later reasoning or planning failure.

\subsection{Additional features}
\label{app:additional_features}
\paragraph{Tutorial room.}
To familiarize agents with the environment’s action space and mitigate confounding variables stemming from procedural uncertainty, we implemented a dedicated, isolated tutorial area at the onset of the simulation. This approach follows established methodological standards in behavioral research and cognitive science \citep{fantz1964visual, saffran1996statistical, schmidt1992new}, as well as empirical game design \citep{andersen2012impact}. This phase ensures that subsequent performance reflects higher-order decision-making rather than a lack of basic control over the game (e.g., action formatting). The tutorial uses a gated, multi-step sequence that requires the successful execution of core interactions, including object manipulation, inventory management, trading, combat, crafting, and navigation. Upon completion, all tutorial-specific assets are removed to ensure a clean transition to the primary experimental trials. Notably, we intentionally omit a subset of available actions in this phase, allowing us to evaluate the agent’s capacity for out-of-distribution learning and the discovery of novel affordances. Both the tutorial and the subsequent quests are integrated into the environment as step rules defined in Section~\ref{sec:ontology}.

\paragraph{Aligning to physical world.}
Despite \abbr{} using LLMs to generate entities and rules, the base game enforces a set of constraints grounded in a physically consistent world. For example, each object has a size attribute, and container objects such as bags have limited capacity. An agent can hold at most two objects simultaneously, corresponding to two hands, and can write notes only when one hand holds a writing tool, such as a pen, and the other holds a writable medium, such as paper. This design choice is motivated by ALFWorld \citep{shridhar2020alfworld}, which aligns a text-based environment with a visual embodied environment and demonstrates that policies learned in an abstract text environment can transfer to more realistic embodied tasks. These constraints introduce explicit trade-offs between holding items, storing them in inventory, or discarding them, thereby encouraging the agent to maintain subgoals, remember object affordances, and manage resources. They also extend the effective horizon of the game: because the agent cannot carry everything at once, it must remember object locations and navigation paths in order to retrieve items later when inventory space becomes available (see Steps 257 and 311 in Figure~\ref{fig:teaser}).

\paragraph{Online expansion.} We also implement an online world expansion option in \abbr{} to ensure that the environment does not run out of content. Implemented as a step rule, it monitors the agent’s visited areas at each step and triggers once all reachable areas have been explored. Upon activation, it performs a difficulty analysis of the current game state and constructs a structured prompt, which is sent asynchronously to an LLM to generate one to two new places, each containing two to three areas, along with level-appropriate objects and NPCs. The generated JSON is then validated and integrated into the world graph: the new areas are connected to the existing graph by connecting them to the furthest reachable area from the spawn location, after which they are populated with both newly generated and existing objects and NPCs.

\paragraph{Respawn and episodic memory.}
Respawn is not a fixed requirement of \abbr{}. The generator LLM or the user can define a step rule that terminates the game after death. Respawning after defeat is also a common design in computer games. All six generated games use respawn because it allows the interaction budget of up to 500 steps to continue after death and creates an episodic memory challenge. The agent must remember the death location to recover its dropped items, unless a generated rule explicitly removes or relocates them. This evaluates episodic memory, one of the five key abilities we propose for test-time continual learning agents. If death terminated the run, evaluating later behavior would require restarting from the beginning and repeating previously completed quest stages. Respawning preserves quest progress and improves evaluation efficiency for long-horizon games.

\begin{figure*}[t]
    \centering
    \includegraphics[width=\textwidth]{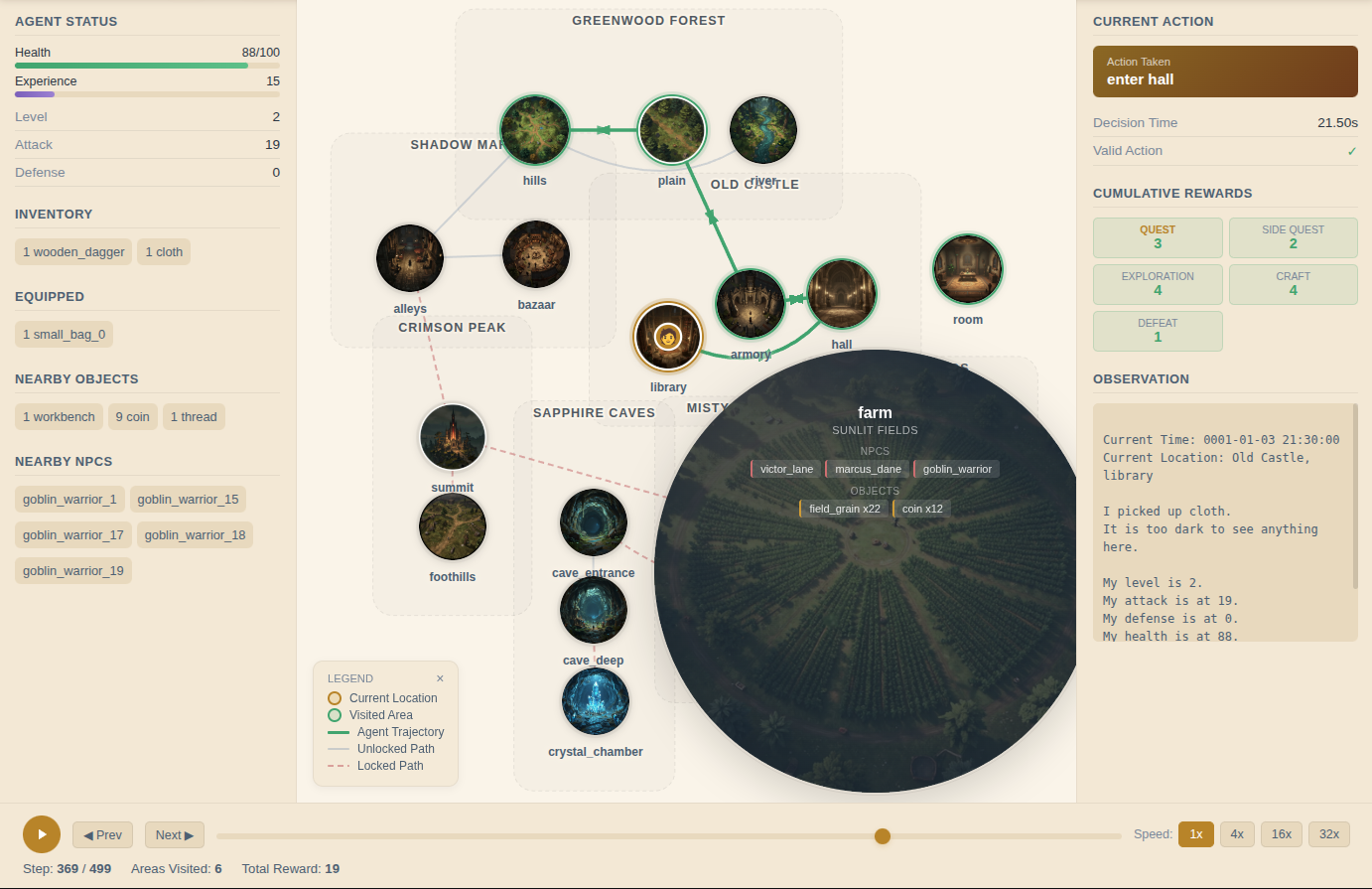}
    \caption{\textbf{Trajectory visualizer.} An interactive web-based tool for replaying agent trajectories over the world graph. The left panel shows agent status; the center displays area nodes with navigable edges, the agent's traversal path, and diffusion-generated icons; and the right panel shows the current action, rewards, and observation text.}
    \label{fig:trajectory_visualizer}
\end{figure*}

\paragraph{Trajectory visualization.}
To support qualitative analysis of agent behavior, we provide an interactive web-based trajectory visualizer that renders the full game state over time. The tool reconstructs the world graph from the environment configuration, laying out areas as circular image nodes grouped by place, with edges indicating navigable connections between them. At each simulation step, the visualizer displays the agent's current location, the action taken, the observation received, the reward breakdown, and a full status panel that includes health, experience, level, inventory, equipped items, and nearby entities (objects and NPCs). A timeline control at the bottom allows the user to play back the trajectory at different speeds or manually scrub to any step, while the agent's traversal path is drawn as a directed overlay on the world graph. Area nodes expand on hover to reveal their current entities, which update dynamically. Node icons are generated automatically using the diffusion model \texttt{Tongyi-MAI/Z-Image-Turbo}, conditioned on entity names and types, producing stylistic image assets for areas, objects, and NPCs. An example visualization is shown in Figure~\ref{fig:trajectory_visualizer}.

\section{Additional agent implementation details}
\label{app:additional_agent_implementation}

We specify the LLM agent hyperparameters used for all agents in Table~\ref{tab:llm_agent_params}, including the agent mechanism short-term memory (STM) size. We also provide additional implementation details below.

\begin{table*}[t]
\hbadness=10000
\setlength{\tabcolsep}{3pt}
\renewcommand{\arraystretch}{1.25}
\centering
\small

\begin{minipage}[t]{0.44\textwidth}
\centering
\caption{\textbf{LLM agent hyper-parameters}}

\begin{tabular}{lc}
\toprule
\textbf{Parameter} & Value \\
\midrule
Temperature             & 0.7 \\
Top P                   & 0.8 \\
Presence Penalty        & 1.5 \\
Max New Tokens          & 4096 \\
STM Size                & 5 \\
\bottomrule
\label{tab:llm_agent_params}
\end{tabular}

\end{minipage}
\hfill
\begin{minipage}[t]{0.55\textwidth}
\centering
\caption{\textbf{Agent mechanisms.} Whether each method has enabled summarization, reflection, or short-term memory.}

\begin{tabular}{lccc}
\toprule
\textbf{Agent Category} & \textbf{SUM} & \textbf{REFL} & \textbf{STM} \\
\midrule

Long Context            & \checkmark & \checkmark & $\times$ \\
Fixed Size              & $\times$   & $\times$   & $\times$ \\
RAG (Vanilla)           & \checkmark & \checkmark & \checkmark \\
SFT                     & \checkmark & \checkmark & \checkmark \\
RL                      & $\times$   & $\times$   & $\times$ \\
Latent                  & $\times$   & $\times$   & $\times$ \\
No Memory               & $\times$   & $\times$   & $\times$ \\
Random                  & $\times$   & $\times$   & $\times$ \\

\bottomrule
\label{tab:agent_mechanisms}
\end{tabular}

\end{minipage}
\end{table*}

\textbf{RAG Agents} use the Qwen3-Embedding-0.6B \citep{zhang2025qwen3} as the embedding model used for retrieval of relevant context. The number of items retrieved at each step is 5 (i.e., top-k).

\textbf{SFT Agents} train the LoRA adapter on the Q, K, V, and O projection matrices, with rank $r = 16$, scaling factor $\alpha = 32$, and dropout rate $0.05$. When short-term memory is disabled, training occurs at every step; when it is enabled, the agent instead trains every five steps on the observation–reasoning–action tuples stored in short-term memory because the short-term memory size is 5 (shown in Table~\ref{tab:llm_agent_params}).

\textbf{No Memory Agents} directly make the decision on the observation, without dependency on memory. 

\textbf{Random Agent} randomly chooses an action from all available actions (with arguments). 

\textbf{RL Agent.} We train an LLM-based agent using reinforcement learning (RL) with the reward function defined in Section~\ref{sec:ontology}. Unlike standard RL setups for LLM agents, which assume episodic environments with state resets, we adapt the training procedure to a test-time continual learning setting. Specifically, we partition the continual interaction stream into episodes while persisting the world state across episodes. Training is performed using PPO \citep{schulman2017proximal} within the verl-agent framework \citep{feng2025group}. We fine-tune Qwen3-4B as the actor and Qwen3-1.7B as the critic. We report results under three configurations. First, we evaluate a minimal online setting with batch size 1 and no short-term memory. Second, we consider a more conventional batch RL setup with a batch size of 16 and parallel environments, evaluated both with and without short-term memory. The episode length equals the short-term memory size. To mitigate the potential advantage of larger batch sizes, for the batch size 16 setting, we report the median performance across independent runs.

\section{Example prompt templates}
\label{app:example_prompt_templates}

\subsection{Agent prompt template}
\label{app:example_agent_prompt}
\begin{lstlisting}[style=mycode]
You are the player in a text adventure game. The world is described in text form. At each turn, you may choose ONE action from the action space below.

Action space:
- attack <npc_name>
- buy <amount> <obj_name> <npc_name>
- craft <amount> <obj_name>
- defend
- disassemble <obj_name>
- discard <amount> <obj_name> <container_name>
- drop <object_name>
- eat <object_name>
- enter <area_name>
- equip <obj_name>
- inspect <object_name>
- lockpick <area_name>
- pick up <object_name>
- pickpocket <npc_name>
- sell <amount> <obj_name> <npc_name>
- store <amount> <obj_name> <container_name>
- take out <obj_name> <container_name>
- talk to <npc_name>
- throw <obj_name> <npc_name>
- unequip <obj_name>
- wait
- write <text> <writable_name>

Output format (STRICT):
Return a single JSON object with exactly these keys:
{{
    "reasoning": "A few sentences explaining why you choose the action.", 
    "action": "<action>"
}}

Rules:
- The JSON must be the ONLY content in your reply (no extra text before/after).
- The action must exactly match one option from the action space.

What should I do next? Return only the JSON object.
\end{lstlisting}

\subsection{Observation}
\label{app:example_obs}

Instance names such as \texttt{cave\_stalker\_6} are canonical identifiers used for unambiguous parsing. Observations can display ``an injured cave stalker'' alongside this identifier, while commands retain the canonical handle. The LLM evaluations use these identifiers for string parsing.

\begin{lstlisting}[style=mycode]
Current Time: 0001-01-01 10:00:00
Current Location: Old Castle, hall

I successfully crafted 1 wood_plank. It is now on the ground.

I am holding 1 torch.
I have equipped 1 small_bag_0.
I see 10 coin, 1 key, 1 paper_7, 1 pen, 1 torch, 2 apple, 1 workbench, 1 wood_plank near me.
I see goblin_warrior_2 nearby.

My level is 1.
My attack is at 10.
My defense is at 0.
My health is at 93.
My experience is at 30.

Neighboring areas: armory, library.
\end{lstlisting}

\section{Examples of synthesized game entities}
\label{app:example_game_entities}
\begin{lstlisting}[style=mycode]
{
  "entities": 
  {
    "places": [
      {
        "type": "place",
        "id": "place_sapphire_caves",
        "name": "Sapphire Caves",
        "unlocked": false,
        "areas": [
          {"type":"area", "id": "area_caves_entrance", "name": "cave_entrance", "level": 2, "light": true},
          {"type":"area", "id": "area_caves_deep", "name": "cave_deep", "level": 4, "light": false},
          {"type":"area", "id": "area_caves_crystal", "name": "crystal_chamber", "level": 5, "light": true}
        ]
      },
      ... (other places)
    ],
    "objects": [
      {
        "type": "object", "id": "obj_meadow_herb", "name": "meadow_herb", "category": "material", 
        "usage": "craft", "value": 2, "size": 1, 
        "description": "A mild herb used for simple healing and warm drinks.", 
        "craft": {"ingredients": {}, "dependencies": []}, "level": 1, 
        "areas": ["area_fields_meadow", "area_forest_plain", "area_fields_grove", "area_caves_entrance"]
      },
      {
        "type": "object", "id": "obj_tension_wrench", "name": "tension_wrench", "category": "tool", 
        "usage": "unlock", "value": 7, "size": 1, 
        "description": "A sturdy wrench to apply torque when picking locks.", 
        "craft": {"ingredients": {"obj_iron_bar": 0.5}, "dependencies": ["obj_workbench"]}, "level": 2
      },
      {
        "type": "object", "id": "obj_lantern", "name": "lantern", "category": "tool", 
        "usage": "light", "value": 30, "size": 3, 
        "description": "A lantern that brightens dark areas when lit.", 
        "craft": {
          "ingredients": {"obj_glass_shard": 2, "obj_sulfur_powder": 1, "obj_cloth_strap": 1}, 
          "dependencies": ["obj_workbench"]
        }, 
        "level": 4
      },
      ... (other objects)
    ],
    "npcs": [
      {
        "type": "npc", "id": "npc_cave_stalker", "name": "cave_stalker", "enemy": true, 
        "unique": false, "role": "beast", 
        "description": "a low, skittering thing that clings to cave walls and lunges from the dark.", 
        "base_attack_power": 6, "base_hp": 40, "slope_hp": 15, "slope_attack_power": 13, 
        "objects": ["obj_quartz_chunk", "obj_cave_salt"], 
        "combat_pattern": ["wait", "attack", "attack", "attack"]
      },
      ... (other npcs)
    ]
  },
  "initializations": {
    "spawn": {
      "area": "area_castle_hall",
      "npcs": {},
      "objects": {
        "obj_coin": 10,
        ... (other objects with quantities)
      }
    }
  },
  "custom_events": ["main_quest", "side_quest", "tutorial"],
  "features": {
    "online_expansion": true,
    "expansion_model": "gpt-5"
  }
}
\end{lstlisting}

\section{Examples of synthesized game dynamics}
\label{app:example_program_synthesized}
We use GPT-5 as the LLM for our generators. We show one synthesized action rule and one synthesized step rule below.

\subsection{Synthesized action rule}
\label{app:example_action_rule}
\begin{lstlisting}[style=mycode]
class EatRule(BaseActionRule):
    name = "action_eat"
    verb = "eat"
    param_min = param_max = 1
    params = ["object_name"]
    description = "Consume a food item to restore HP."

    def apply(self, ctx: RuleContext, res: RuleResult) -> None:
        env, world, agent = ctx.env, ctx.world, ctx.agent
        obj_name = ctx.params[0]

        if obj_name not in world.auxiliary["obj_name_to_id"]:
            res.add_feedback(agent.id, f"Cannot eat {obj_name},"
                     " not found in hand, inventory, held containers, or on the ground.\n")"
            return

        . . . (a series of validation checks)

        current_area = world.area_instances[env.curr_agents_state["area"][agent.id]]
        # determine source: hand -> inventory -> held containers -> ground
        src_loc = None
        consumed_oid = None
        from_label = None

        if obj_id in agent.items_in_hands and agent.items_in_hands[obj_id] > 0:
            agent.items_in_hands[obj_id] -= 1
            if agent.items_in_hands[obj_id] == 0:
                del agent.items_in_hands[obj_id]
            src_loc = res.tloc("hand", agent.id)
            consumed_oid = obj_id
            from_label = "hand"

        elif . . .
    
        hp_restore = int(getattr(obj_def, "hp_restore", 0))
        restore_amount = hp_restore if hp_restore > 0 else 10
        prev_hp = agent.hp
        agent.hp = min(agent.max_hp, agent.hp + restore_amount)
        restored = agent.hp - prev_hp

        res.track_consume(agent.id, consumed_oid, 1, src=src_loc)
        res.add_feedback(...)
        res.events.append(Event(type="object_consumed"...))
\end{lstlisting}

\subsection{Synthesized step rule}
\label{app:example_step_rule}
\begin{lstlisting}[style=mycode]
class ContinuousCraftingMomentumStepRule(BaseStepRule):
    name = "continuous_crafting_momentum_step"
    description = "Awards coin refunds when an agent crafts in the same area on consecutive steps."
    priority = 5

    def __init__(self) -> None:
        super().__init__()
        self._last_step_range = 5  # how many steps back to consider for consecutive crafting
        self._max_coins = 2  # maximum possible coins refunded per crafted item

    def apply(self, ctx: RuleContext, res: RuleResult) -> None:
        env, world = ctx.env, ctx.world
        if env is None or world is None:
            return

        ... (if at tutorial room, skip)

        env.curr_agents_state.setdefault("craft_streak", {})

        current_step = int(env.steps)
        for agent in env.agents:
            # collect all crafting events for this agent this step
            crafted_events = [
                e for e in res.events
                if getattr(e, "agent_id", None) == agent.id and getattr(e, "type", None) == "object_crafted"
            ]
            if not crafted_events:
                continue

            total_crafted = 0
            for ev in crafted_events:
                try:
                    total_crafted += int((getattr(ev, "data", {}) or {}).get("amount", 1))
                except Exception:
                    total_crafted += 1

            area_id = env.curr_agents_state["area"][agent.id]
            record = env.curr_agents_state["craft_streak"].get(agent.id, {
                "last_step": None, "last_area": None, "streak": 0
            })

            if record.get("last_step") is None:
                consecutive = False
            else:
                consecutive = (record.get("last_step") >= current_step - self._last_step_range) and (record.get("last_area") == area_id)
            streak = int(record.get("streak", 0)) + 1 if consecutive else 1

            # reward kicks in starting from the second consecutive craft in the same area.
            if streak >= 2 and total_crafted > 0:
                bonus_coins = total_crafted * env.rng.randint(1, self._max_coins)
                coin_id = "obj_coin"

                area = world.area_instances.get(area_id)
                if area is not None:
                    area.objects[coin_id] = int(area.objects.get(coin_id, 0)) + bonus_coins
                    res.track_spawn("env", coin_id, bonus_coins, res.tloc("area", area_id))
                    res.add_feedback(...)
                    res.events.append(Event(type="craft_streak_bonus" ... ))

            env.curr_agents_state["craft_streak"][agent.id] = {
                "last_step": current_step,
                "last_area": area_id,
                "streak": streak,
            }
\end{lstlisting}

\section{Generated game details}
\label{app:generated_game_details}
We use the game generator described in Section~\ref{sec:game_generation} with GPT-5 \citep{singh2025openai} to generate all six games in Table~\ref{tab:game_overview}. Remnant and Mark are used for the detailed studies in Sections~\ref{sec:case_study} and~\ref{sec:agent_mechanisms}. The four other games were generated to demonstrate diversity and are also evaluated in Section~\ref{sec:multi_game}. Tables~\ref{tab:game_1_stats},~\ref{tab:game_2_stats}, and~\ref{tab:4_game_stats} summarize their entities, rules, and diagnostic questions.

Across six games, scale ranges from 4--18 areas, 35--101 object types, 17--23 action rule types, and 8--28 main quest stages. Their dynamics include entry fees, movement of ground objects, power failures, and persistent forge heat. Their quests require different combinations of evidence, routing, resource recovery, combat, and timing. The generator can produce games with as many entities, action rules, and step rules as desired, while each game is finite. Grounded commands vary with state, so we report action rule types rather than grounded command counts. Five games provide four side quests per area. Mark intentionally has none so that we can study test-time training without goal switching between main and side quests.

\begin{table}[ht]
\centering
\small
\caption{\textbf{Overview of the six generated games.}}
\label{tab:game_overview}
\begin{tabularx}{\textwidth}{lXr}
\toprule
Game & Main quest & Stages \\
\midrule
Remnant & Forge the Confluence Key and defeat the Engine's guardian. & 24 \\
Mark & Obtain the Tear of Forest and defeat its guardian. & 17 \\
Metropolis & Defeat Magistrate Vosk and restore the city charter. & 21 \\
Saltglass & Learn the surveyor's oath and deliver a certified chart. & 8 \\
Quarantine & Synthesize an antivirus, defeat the Tyrant, and deploy the cure. & 28 \\
Robot Kingdom & Destroy the Infernal Architect and end the siege. & 21 \\
\bottomrule
\end{tabularx}
\end{table}

\begin{table*}[t]
\hbadness=10000
\setlength{\tabcolsep}{3pt}
\renewcommand{\arraystretch}{1.25}
\centering
\small

\begin{minipage}[t]{0.49\textwidth}
\centering
\caption{\textbf{Experiment 1 game statistics}}
\label{tab:game_1_stats}
\begin{tabularx}{\linewidth}{@{}p{0.18\linewidth}p{0.15\linewidth}X@{}}
\toprule
\textbf{Type} & \textbf{Count} & \textbf{Example} \\
\midrule
Area & 18 & Old Castle, hall \\
Object & 83 types & wood log, mushroom stew, workbench, wooden sword \\
NPC & 13 types & goblin warrior, villager, ethan\_park \\
Act. Rules & 22 & store, drop, eat, attack, defend, lockpick, inspect \\
Step Rules & 18 & Combat pattern: NPCs vary in attack, defense, and waiting; agents must adapt to win. \\
Main Quest & 24 stages & Defeat cinder\_reaver, the guardian that rises from the shrine’s ash. \\
Side Quest & 4 / area & Collect 5 threads. \\
Know. QA & 116 & What ingredient is needed to craft the object lockpick? \\
Epi. QA & varies & Where did you last drop the iron sword? \\
\bottomrule
\end{tabularx}

\end{minipage}
\hfill
\begin{minipage}[t]{0.49\textwidth}
\centering
\caption{\textbf{Experiment 2 game statistics}}
\label{tab:game_2_stats}
\begin{tabularx}{\linewidth}{@{}p{0.18\linewidth}p{0.15\linewidth}X@{}}
\toprule
\textbf{Type} & \textbf{Count} & \textbf{Example} \\
\midrule
Areas & 14 & Night Market, stalls \\
Object & 49 types & oak rod, goblin sword, lantern, leather strip \\
NPC & 12 types & cave chitter, commoner, maya\_wells \\
Act. Rules & 21 & throw, take out, sell, pickpocket, talk to \\
Step Rules & 17 & Rumor mill: After notable events, merchants may leave scribbled rumor notes that can grant tips or hush noisy areas. \\
Main Quest & 17 stages & Trade with the librarian and obtain the Tear of Forest. \\
Side Quest & -- & -- \\
Know. QA & 110 & What area is connected to the area farm? \\
Epi. QA & varies & Which object did you craft? \\
\bottomrule
\end{tabularx}

\end{minipage}
\end{table*}

\begin{table*}[t]
\hbadness=10000
\setlength{\tabcolsep}{3pt}
\renewcommand{\arraystretch}{1.25}
\centering
\small

\caption{\textbf{Game statistics of the other 4 generated games.}}
\label{tab:4_game_stats}

\begin{minipage}[t]{0.49\textwidth}
\centering
\textbf{Metropolis}\\[2pt]
\begin{tabularx}{\linewidth}{@{}p{0.18\linewidth}p{0.15\linewidth}X@{}}
\toprule
\textbf{Type} & \textbf{Count} & \textbf{Example} \\
\midrule
Areas & 15 & Verdict Spire, gallery \\
Object & 64 types & gavelwood splinter, briefcase, verdict blade, ink \\
NPC & 8 types & bailiff mender, public defender, verdict broker \\
Act. Rules & 17 & salvage, pickpocket, peek \\
Step Rules & 12 & Custodary fee: A small coin/debuff tax is deducted from the agent's inventory when entering a governed area. \\
Main Quest & 21 stages & Demonstrate readiness: possess any emblem of judgment or defense. \\
Side Quest & 4 per area & Talk to public defender. \\
\bottomrule
\end{tabularx}
\end{minipage}
\hfill
\begin{minipage}[t]{0.49\textwidth}
\centering
\textbf{Robot Kingdom}\\[2pt]
\begin{tabularx}{\linewidth}{@{}p{0.18\linewidth}p{0.15\linewidth}X@{}}
\toprule
\textbf{Type} & \textbf{Count} & \textbf{Example} \\
\midrule
Areas & 8 & Ashen Ramparts, gate breach \\
Object & 73 types & machine breaker, coil spear, machine schematic \\
NPC & 4 types & aria coilbinder, ashen harvester, lady viera broken banner \\
Act. Rules & 23 & set trap, smoke, jam, consecrate ward \\
Step Rules & 18 & Forge heat refinement: After a devil-forged machine is slain, the area retains forge-heat that hurts agents at each step for a short time. \\
Main Quest & 21 stages & Test a resonance sweep where machines gather. \\
Side Quest & 4 per area & Trade 1 hell rivet with ashen harvester. \\
\bottomrule
\end{tabularx}
\end{minipage}

\vspace{0.8em}

\begin{minipage}[t]{0.49\textwidth}
\centering
\textbf{Quarantine}\\[2pt]
\begin{tabularx}{\linewidth}{@{}p{0.18\linewidth}p{0.15\linewidth}X@{}}
\toprule
\textbf{Type} & \textbf{Count} & \textbf{Example} \\
\midrule
Areas & 12 & West Sewer Junction, flooded tunnels \\
Object & 101 types & chem fluid, gun power, electronics kit \\
Act. Rules & 18 & barricade path, suppress noise, search \\
Step Rules & 18 & Power grid failure: At random intervals, certain areas will lose power -- disabling crafting stations and losing visibility. \\
Main Quest & 28 stages & Prove you can survive by defeating a threat. \\
Side Quest & 4 per area & Craft 1 kevlar vest. \\
\bottomrule
\end{tabularx}
\end{minipage}
\hfill
\begin{minipage}[t]{0.49\textwidth}
\centering
\textbf{Saltglass}\\[2pt]
\begin{tabularx}{\linewidth}{@{}p{0.18\linewidth}p{0.15\linewidth}X@{}}
\toprule
\textbf{Type} & \textbf{Count} & \textbf{Example} \\
\midrule
Areas & 4 & Saltglass Expanse, saints waystation \\
Object & 35 types & memory splinter, annealed saltglass, noonstorm core \\
Act. Rules & 19 & placate, chart route \\
Step Rules & 14 & Salt gust shuffle: Each step spent in a sunlit, wind-exposed area, fickle salt gusts may blow one small ground item through an unlocked path into a neighboring area. \\
Main Quest & 8 stages & Bring the lens to the mirage port and hold it up until the way resolves. \\
Side Quest & 4 per area & Collect 3 brine mote. \\
\bottomrule
\end{tabularx}
\end{minipage}

\end{table*}

\clearpage
\input{sections/generator_evaluation}

%% file: sections/results_table.tex
\section{More results for Experiment 1 and Experiment 2}
\label{app:more_results}

\begin{table}[H]
\centering
\caption{\textbf{Experiment 1 results with proprietary LLM backbones.} $\uparrow$ and $\downarrow$ indicate that higher and lower values are better, respectively. Best results in each column are highlighted in bold. \textbf{Q} denotes the main quest progress reward. \textbf{SQ}, \textbf{E}, \textbf{C}, and \textbf{D} denote supplementary rewards for side quests, area exploration, crafting, and defeating, respectively. \textbf{WK.b} and \textbf{WK.a} denote World Knowledge QA accuracy before and after gameplay, respectively, while \textbf{Epi.a} denotes the Episodic Memory QA accuracy after gameplay. \textbf{OE} and \textbf{AE} denote object and action exploration counts, respectively. \textbf{AD}, \textbf{AT}, and \textbf{IA} denote action diversity, average token usage per step, and invalid-action rate, respectively.}
\scriptsize
\setlength{\tabcolsep}{2pt}
\renewcommand{\arraystretch}{1}
\resizebox{\textwidth}{!}{%
\begin{tabular}{@{}lllccccccccccccc@{}}
\toprule
& & & \multicolumn{1}{c}{\textit{Main}} & \multicolumn{4}{c}{\textit{Supplementary}} & & & & & & & & \\
\textbf{Category} & \textbf{LLM} & \textbf{Method} &
\textbf{Q$\uparrow$} &
\textbf{SQ$\uparrow$} & \textbf{E$\uparrow$} & \textbf{C$\uparrow$} & \textbf{D$\uparrow$} &
\multicolumn{1}{c}{\textbf{WK.b}} & \multicolumn{1}{c}{\textbf{WK.a$\uparrow$}} &
\multicolumn{1}{c}{\textbf{Epi.a$\uparrow$}} &
\multicolumn{1}{c}{\textbf{OE$\uparrow$}} & \multicolumn{1}{c}{\textbf{AE$\uparrow$}} &
\multicolumn{1}{c}{\textbf{AD$\uparrow$}} &
\multicolumn{1}{c}{\textbf{AT$\downarrow$}} &
\multicolumn{1}{c}{\textbf{IA$\downarrow$}} \\
\midrule
\rowcolor{gray!20}
Human & -- & -- & 9 & 8 & 10 & 11 & 11 & -- & -- & -- & -- & -- & -- & -- & -- \\
\midrule
RAG & GPT-5 & Baseline & 1 & 0 & 2 & 0 & 2 & 0.322 & 0.487 & 0.709 & 6 / 83 & 12 / 22 & 0.311 & 2051.3 & \textbf{0.00} \\
RAG & GPT-5 & Reflection & 1 & 0 & 2 & 0 & 0 & 0.296 & 0.409 & 0.729 & 0 / 83 & 8 / 22 & 0.101 & 6943.6 & \textbf{0.00} \\
RAG & GPT-5 & Summarization & 1 & 0 & 2 & 0 & 0 & 0.339 & 0.504 & 0.720 & 3 / 83 & 10 / 22 & 0.149 & 2210.2 & \textbf{0.00} \\
RAG & GPT-5 & STM & 2 & 2 & 2 & 3 & \textbf{8} & 0.348 & 0.565 & 0.465 & 13 / 83 & 15 / 22 & 0.696 & 3380.2 & \textbf{0.00} \\
RAG & GPT-5 & Raptor & 1 & 0 & 2 & 0 & 3 & 0.330 & 0.443 & 0.436 & 4 / 83 & 11 / 22 & 0.326 & 1540.8 & \textbf{0.00} \\
RAG & GPT-5 & Mem0 & 0 & 0 & 1 & 0 & 3 & 0.313 & 0.313 & 0.233 & 8 / 83 & 11 / 22 & 0.559 & 1114.9 & 0.01 \\
RAG & GPT-5 & Voyager & 1 & 1 & 2 & 1 & 7 & 0.339 & 0.400 & 0.427 & 10 / 83 & 13 / 22 & 0.676 & 1384.6 & 0.03 \\
\midrule
RAG & GPT-5-mini & Baseline & 1 & 0 & 2 & 0 & \textbf{8} & 0.226 & 0.435 & 0.473 & 6 / 83 & 13 / 22 & 0.675 & 1991.9 & \textbf{0.00} \\
RAG & GPT-5-mini & Reflection & 1 & 0 & 1 & 0 & 0 & 0.261 & 0.409 & 0.633 & 3 / 83 & 8 / 22 & 0.311 & 6729.1 & \textbf{0.00} \\
RAG & GPT-5-mini & Summarization & 1 & 0 & 2 & 0 & 3 & 0.322 & 0.461 & 0.527 & 5 / 83 & 11 / 22 & 0.489 & 2245.0 & 0.02 \\
RAG & GPT-5-mini & STM & 1 & 0 & 2 & 0 & 4 & 0.278 & 0.565 & 0.453 & 9 / 83 & 14 / 22 & 0.682 & 3021.7 & 0.01 \\
RAG & GPT-5-mini & Raptor & 1 & 0 & 1 & 0 & 5 & 0.252 & 0.409 & 0.360 & 8 / 83 & 13 / 22 & 0.655 & 1390.8 & \textbf{0.00} \\
RAG & GPT-5-mini & Mem0 & 1 & 0 & 1 & 1 & 4 & 0.261 & 0.270 & 0.236 & 12 / 83 & 14 / 22 & 0.655 & \textbf{874.0} & \textbf{0.00} \\
RAG & GPT-5-mini & Voyager & 1 & 2 & 2 & 0 & 0 & 0.261 & 0.339 & 0.500 & 7 / 83 & 14 / 22 & 0.658 & 1196.6 & 0.04 \\
\midrule
Long Context & GPT-5 & Baseline & \textbf{3} & 2 & \textbf{4} & \textbf{6} & 1 & 0.365 & \textbf{0.713} & 0.918 & \textbf{18 / 83} & \textbf{17 / 22} & 0.631 & 51856.8 & \textbf{0.00} \\
Long Context & GPT-5 & Reflection & 2 & 2 & 2 & 3 & 0 & 0.304 & 0.635 & 0.766 & 13 / 83 & 16 / 22 & 0.617 & 125239.2 & \textbf{0.00} \\
Long Context & GPT-5 & Summarization & 2 & \textbf{3} & 2 & \textbf{6} & \textbf{8} & 0.322 & 0.617 & \textbf{0.922} & \textbf{18 / 83} & 16 / 22 & 0.605 & 36849.7 & \textbf{0.00} \\
\midrule
Long Context & GPT-5-mini & Baseline & 2 & \textbf{3} & 2 & 3 & 4 & 0.270 & 0.609 & 0.843 & 14 / 83 & 15 / 22 & 0.680 & 60783.7 & \textbf{0.00} \\
Long Context & GPT-5-mini & Reflection & 1 & 2 & 2 & 0 & 0 & 0.252 & 0.443 & 0.564 & 4 / 83 & 14 / 22 & 0.689 & 134010.6 & \textbf{0.00} \\
Long Context & GPT-5-mini & Summarization & 1 & 2 & 2 & 0 & 4 & 0.287 & 0.522 & 0.877 & 7 / 83 & 14 / 22 & 0.702 & 47510.4 & \textbf{0.00} \\
\midrule
Fixed Size & GPT-5 & STM & 1 & 2 & 2 & 0 & 5 & 0.313 & 0.409 & 0.587 & 9 / 83 & 16 / 22 & 0.628 & 2362.8 & 0.01 \\
Fixed Size & GPT-5 & MEM1 & 0 & 0 & 1 & 0 & 4 & 0.365 & 0.400 & 0.380 & 12 / 83 & \textbf{17 / 22} & 0.696 & 2259.4 & \textbf{0.00} \\
\midrule
Fixed Size & GPT-5-mini & STM & 1 & 0 & 2 & 0 & 2 & 0.235 & 0.313 & 0.323 & 13 / 83 & \textbf{17 / 22} & 0.686 & 1828.0 & 0.01 \\
Fixed Size & GPT-5-mini & MEM1 & 1 & 0 & 1 & 0 & 2 & 0.287 & 0.365 & 0.273 & 4 / 83 & 13 / 22 & \textbf{0.709} & 1746.5 & \textbf{0.00} \\
\midrule
No Memory & GPT-5 & -- & 0 & 0 & 2 & 0 & 2 & 0.339 & 0.322 & 0.208 & 2 / 83 & 10 / 22 & 0.414 & 1062.4 & 0.02 \\
\midrule
No Memory & GPT-5-mini & -- & 1 & 0 & 1 & 0 & 4 & 0.270 & 0.243 & 0.197 & 10 / 83 & 14 / 22 & 0.639 & 896.5 & 0.02 \\
\midrule
Random & -- & -- & -- & -- & -- & -- & -- & -- & -- & -- & -- & -- & -- & -- & -- \\

\bottomrule
\end{tabular}%
}
\label{tab:results_proprietary_exp1}
\end{table}

\begin{center}
    \captionsetup{hypcap=false}
    \begin{minipage}[b]{0.70\linewidth}
        \centering
        \includegraphics[width=\linewidth]{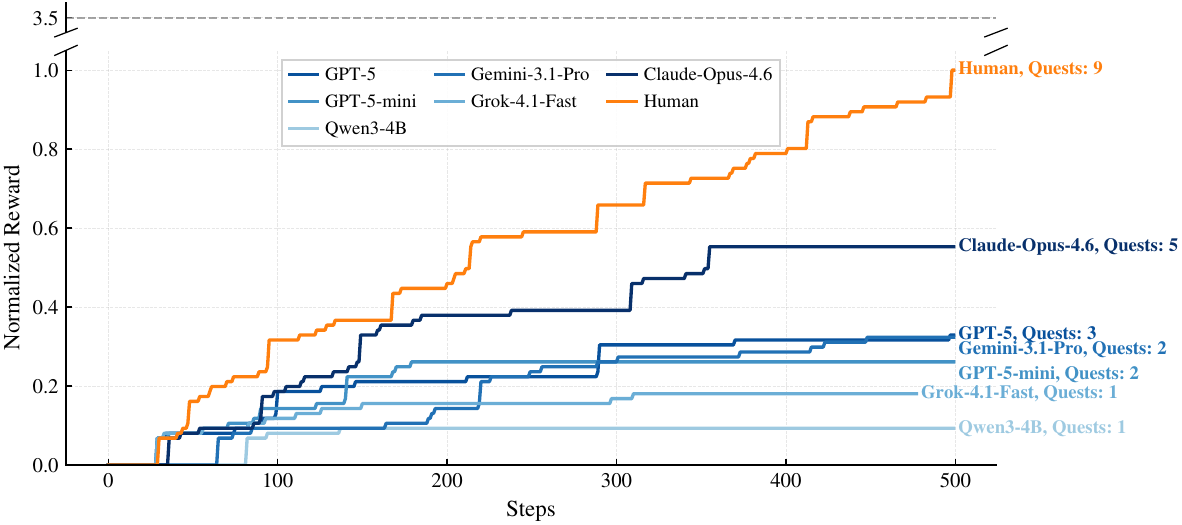}
        \captionof{figure}{Cumulative reward of more frontier LLMs using the Long Context agent, compared with human performance in the Experiment 1 game.}
        \label{fig:cumulative_reward_lc_all_llms}
    \end{minipage}
    \hfill
    \begin{minipage}[b]{0.27\linewidth}
        \centering
        \includegraphics[width=\linewidth]{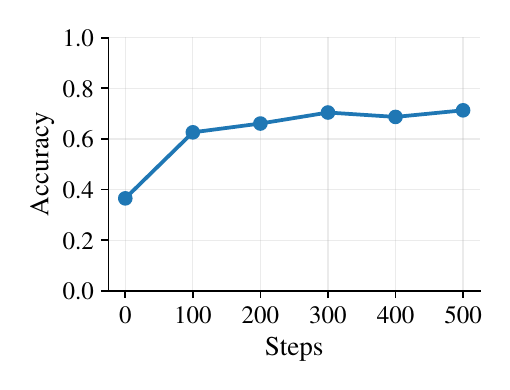}
        \captionof{figure}{World Knowledge QA every 100 steps.}
        \label{fig:progressive_wk_qa}
    \end{minipage}
\end{center}
\vspace{-14pt}

\begin{table}[H]
\centering
\caption{\textbf{Experiment 1 results with open-weight LLM backbones.} $\uparrow$ and $\downarrow$ indicate that higher and lower values are better, respectively. Best results in each column are highlighted in bold. \textbf{Q} denotes the main quest progress reward. \textbf{SQ}, \textbf{E}, \textbf{C}, and \textbf{D} denote supplementary rewards for side quests, area exploration, crafting, and defeating, respectively. \textbf{WK.b} and \textbf{WK.a} denote World Knowledge QA accuracy before and after gameplay, respectively, while \textbf{Epi.a} denotes the Episodic Memory QA accuracy after gameplay. \textbf{OE} and \textbf{AE} denote object and action exploration counts, respectively. \textbf{AD}, \textbf{AT}, and \textbf{IA} denote action diversity, average token usage per step, and invalid-action rate, respectively. \textit{Because MemoryLLM and MPlus never produce valid actions, we opt out of testing episodic memory, since the episodic experience contains only the default wait action.}}
\scriptsize
\setlength{\tabcolsep}{2pt}
\renewcommand{\arraystretch}{1}
\resizebox{\textwidth}{!}{%
\begin{tabular}{@{}lllccccccccccccc@{}}
\toprule
& & & \multicolumn{1}{c}{\textit{Main}} & \multicolumn{4}{c}{\textit{Supplementary}} & & & & & & & & \\
\textbf{Category} & \textbf{LLM} & \textbf{Method} &
\textbf{Q$\uparrow$} &
\textbf{SQ$\uparrow$} & \textbf{E$\uparrow$} & \textbf{C$\uparrow$} & \textbf{D$\uparrow$} &
\multicolumn{1}{c}{\textbf{WK.b}} & \multicolumn{1}{c}{\textbf{WK.a$\uparrow$}} &
\multicolumn{1}{c}{\textbf{Epi.a$\uparrow$}} &
\multicolumn{1}{c}{\textbf{OE$\uparrow$}} & \multicolumn{1}{c}{\textbf{AE$\uparrow$}} &
\multicolumn{1}{c}{\textbf{AD$\uparrow$}} &
\multicolumn{1}{c}{\textbf{AT$\downarrow$}} &
\multicolumn{1}{c}{\textbf{IA$\downarrow$}} \\
\midrule
\rowcolor{gray!20}
Human & -- & -- & 9 & 8 & 10 & 11 & 11 & -- & -- & -- & -- & -- & -- & -- & -- \\
\midrule
RAG & Qwen3-4B & Baseline & \textbf{1} & 0 & \textbf{2} & 0 & 1 & 0.235 & 0.409 & 0.312 & 0 / 83 & 17 / 22 & 0.547 & 1890.7 & 0.03 \\
RAG & Qwen3-4B & Reflection & 0 & 0 & 1 & 0 & 0 & 0.235 & 0.426 & 0.306 & 0 / 83 & 15 / 22 & 0.684 & 2725.9 & 0.17 \\
RAG & Qwen3-4B & Summarization & 0 & 0 & 1 & 0 & 0 & 0.235 & 0.374 & 0.396 & 0 / 83 & 12 / 22 & 0.452 & 2121.8 & \textbf{0.01} \\
RAG & Qwen3-4B & STM & \textbf{1} & \textbf{2} & \textbf{2} & 0 & 2 & 0.235 & \textbf{0.452} & 0.309 & 3 / 83 & 17 / 22 & 0.721 & 3175.6 & 0.04 \\
RAG & Qwen3-4B & Raptor & 0 & 0 & 1 & 0 & 3 & 0.235 & 0.426 & 0.327 & 2 / 83 & 16 / 22 & 0.684 & 1530.4 & 0.05 \\
RAG & Qwen3-4B & Mem0 & \textbf{1} & 1 & \textbf{2} & 0 & 6 & 0.252 & 0.252 & 0.151 & 7 / 83 & 13 / 22 & 0.712 & 1241.5 & 0.10 \\
RAG & Qwen3-4B & Voyager & \textbf{1} & \textbf{2} & \textbf{2} & 0 & 2 & 0.235 & 0.348 & 0.299 & 7 / 83 & \textbf{20 / 22} & \textbf{0.801} & 1248.8 & 0.12 \\
\midrule
Long Context & Qwen3-4B & Baseline & \textbf{1} & 0 & 1 & 0 & 2 & 0.235 & 0.000 & 0.000 & 0 / 83 & 14 / 22 & 0.557 & 64097.8 & 0.28 \\
Long Context & Qwen3-4B & Reflection & \textbf{1} & 0 & \textbf{2} & 0 & \textbf{7} & 0.235 & 0.000 & 0.000 & 1 / 83 & 12 / 22 & 0.550 & 73875.6 & 0.35 \\
Long Context & Qwen3-4B & Summarization & 0 & 0 & 0 & 0 & 0 & 0.235 & 0.339 & 0.391 & 1 / 83 & 7 / 22 & 0.343 & 15417.9 & 0.04 \\
\midrule
Fixed Size & Qwen3-4B & STM & 0 & 0 & 1 & \textbf{1} & 1 & 0.235 & 0.391 & 0.271 & 6 / 83 & 18 / 22 & 0.759 & 2564.5 & 0.08 \\
Fixed Size & Qwen3-4B & MEM1 & 0 & 0 & 0 & 0 & 0 & 0.235 & 0.330 & 0.065 & 1 / 83 & 8 / 22 & 0.552 & 1324.7 & 0.20 \\
\midrule
SFT & Qwen3-4B & Baseline & 0 & 0 & 0 & 0 & 0 & 0.235 & 0.235 & 0.130 & 1 / 83 & 10 / 22 & 0.470 & 993.9 & 0.06 \\
SFT & Qwen3-4B & Reflection & 0 & 0 & 0 & 0 & 0 & 0.235 & 0.096 & 0.156 & 0 / 83 & 0 / 22 & 0.000 & 2339.9 & 0.36 \\
SFT & Qwen3-4B & Summarization & \textbf{1} & 0 & \textbf{2} & 0 & 1 & 0.235 & 0.122 & 0.104 & 0 / 83 & 13 / 22 & 0.491 & 1276.7 & 0.25 \\
SFT & Qwen3-4B & STM & \textbf{1} & 0 & \textbf{2} & 0 & 4 & 0.235 & 0.104 & 0.050 & 1 / 83 & 19 / 22 & 0.732 & 2442.0 & 0.21 \\
\midrule
RL & Qwen3-4B & Baseline & 0 & 0 & 0 & 0 & 0 & 0.235 & 0.157 & 0.109 & 1 / 83 & 9 / 22 & 0.465 & 1098.3 & 0.06 \\
RL & Qwen3-4B & 16 envs & \textbf{1} & 1 & \textbf{2} & 0 & \textbf{7} & 0.235 & 0.217 & 0.151 & 6 / 83 & 18 / 22 & 0.719 & 1288.2 & 0.06 \\
RL & Qwen3-4B & STM (16 envs) & \textbf{1} & 1 & \textbf{2} & 0 & 6 & 0.235 & 0.226 & 0.183 & \textbf{9 / 83} & 19 / 22 & 0.777 & 2307.6 & 0.08 \\
\midrule
Latent & LLaMA3-8B & MemoryLLM & 0 & 0 & 0 & 0 & 0 & 0.104 & 0.191 & -- & 0 / 83 & 0 / 22 & 0.000 & 771.1 & 1.00 \\
\midrule
Latent & LLaMA3.1-8B & MPlus & 0 & 0 & 0 & 0 & 0 & 0.000 & 0.043 & -- & 0 / 83 & 0 / 22 & 0.000 & \textbf{476.9} & 1.00 \\
\midrule
No Memory & Qwen3-4B & -- & \textbf{1} & 1 & \textbf{2} & 0 & 1 & 0.235 & 0.235 & 0.196 & 3 / 83 & 14 / 22 & 0.647 & 1011.0 & 0.10 \\
\midrule
Random & -- & -- & -- & -- & -- & -- & -- & -- & -- & -- & -- & -- & -- & -- & -- \\

\bottomrule
\end{tabular}%
}
\label{tab:results_open_exp1}
\end{table}

\vspace{-15pt}

\begin{table}[H]
\centering
\caption{\textbf{Experiment 2 results with Qwen3-4B.} $\uparrow$ and $\downarrow$ indicate that higher and lower values are better, respectively. Best results in each column are highlighted in bold. \textbf{Q} denotes the main quest progress reward. \textbf{SQ}, \textbf{E}, \textbf{C}, and \textbf{D} denote supplementary rewards for side quests, area exploration, crafting, and defeating, respectively. \textbf{WK.b} and \textbf{WK.a} denote World Knowledge QA accuracy before and after gameplay, respectively, while \textbf{Epi.a} denotes the Episodic Memory QA accuracy after gameplay. \textbf{OE} and \textbf{AE} denote object and action exploration counts, respectively. \textbf{AD}, \textbf{AT}, and \textbf{IA} denote action diversity, average token usage per step, and invalid-action rate, respectively.}
\scriptsize
\setlength{\tabcolsep}{2pt}
\renewcommand{\arraystretch}{1}
\resizebox{\textwidth}{!}{%
\begin{tabular}{@{}lllccccccccccccc@{}}
\toprule
& & & \multicolumn{1}{c}{\textit{Main}} & \multicolumn{4}{c}{\textit{Supplementary}} & & & & & & & & \\
\textbf{Category} & \textbf{LLM} & \textbf{Method} &
\textbf{Q$\uparrow$} &
\textbf{SQ$\uparrow$} & \textbf{E$\uparrow$} & \textbf{C$\uparrow$} & \textbf{D$\uparrow$} &
\multicolumn{1}{c}{\textbf{WK.b}} & \multicolumn{1}{c}{\textbf{WK.a$\uparrow$}} &
\multicolumn{1}{c}{\textbf{Epi.a$\uparrow$}} &
\multicolumn{1}{c}{\textbf{OE$\uparrow$}} & \multicolumn{1}{c}{\textbf{AE$\uparrow$}} &
\multicolumn{1}{c}{\textbf{AD$\uparrow$}} &
\multicolumn{1}{c}{\textbf{AT$\downarrow$}} &
\multicolumn{1}{c}{\textbf{IA$\downarrow$}} \\
\midrule
\rowcolor{gray!20}
Human & -- & -- & 17 & -- & 10 & 10 & 8 & -- & -- & -- & -- & -- & -- & -- & -- \\
\midrule
SFT & Qwen3-4B & Baseline & 0 & -- & 2 & 0 & 3 & 0.091 & 0.036 & 0.016 & 4 / 48 & 16 / 21 & 0.747 & 1077.5 & 0.08 \\
SFT & Qwen3-4B & Reflection & 0 & -- & 1 & 0 & 0 & 0.091 & 0.018 & 0.000 & 2 / 48 & 16 / 21 & 0.609 & 1977.3 & 0.46 \\
SFT & Qwen3-4B & Summarization & 1 & -- & 1 & 1 & 2 & 0.091 & 0.027 & 0.053 & 3 / 48 & 16 / 21 & 0.692 & \textbf{1069.9} & 0.20 \\
SFT & Qwen3-4B & STM & \textbf{7} & -- & \textbf{10} & 4 & 2 & 0.091 & 0.036 & 0.051 & 9 / 48 & 18 / 21 & \textbf{0.777} & 2436.6 & 0.13 \\
\midrule
Long Context & Qwen3-4B & Baseline & 0 & -- & 1 & 1 & 2 & 0.091 & 0.000 & 0.000 & 4 / 48 & 12 / 21 & 0.542 & 66811.8 & 0.32 \\
\midrule
RAG & Qwen3-4B & Baseline & 1 & -- & 2 & 2 & \textbf{7} & 0.091 & \textbf{0.336} & \textbf{0.304} & 6 / 48 & 17 / 21 & 0.659 & 2205.3 & \textbf{0.01} \\
\midrule
Fixed Size & Qwen3-4B & STM & 6 & -- & 2 & \textbf{5} & 3 & 0.091 & 0.264 & 0.221 & \textbf{11 / 48} & \textbf{20 / 21} & 0.762 & 2284.5 & 0.03 \\

\bottomrule
\end{tabular}%
}
\label{tab:results_exp2}
\end{table}

\begin{table}[H]
\centering
\caption{\textbf{Effect of short-term memory size on agent performance.} $\uparrow$ and $\downarrow$ indicate that higher and lower values are better, respectively. \textbf{Q} denotes the main quest progress reward. \textbf{SQ}, \textbf{E}, \textbf{C}, and \textbf{D} denote supplementary rewards for side quests, area exploration, crafting, and defeating, respectively. \textbf{WK.a} denotes the World Knowledge QA accuracy after gameplay, and \textbf{Epi.a} denotes the Episodic Memory QA accuracy after gameplay. \textbf{OE} and \textbf{AE} denote object and action exploration counts, respectively. \textbf{AD}, \textbf{AT}, and \textbf{IA} denote action diversity, average token usage per step, and invalid-action rate, respectively. \textit{The short-term memory size of 0 corresponds to the No Memory baseline, while size 500 corresponds to the Long Context agent.}}
\scriptsize
\setlength{\tabcolsep}{2pt}
\renewcommand{\arraystretch}{1}
\resizebox{\textwidth}{!}{%
\begin{tabular}{@{}lllccccccccccccc@{}}
\toprule
& & & & \multicolumn{1}{c}{\textit{Main}} & \multicolumn{4}{c}{\textit{Supplementary}} & & & & & & & \\
\textbf{Category} & \textbf{LLM} & \textbf{Method} & \textbf{STM Size} &
\textbf{Q$\uparrow$} &
\textbf{SQ$\uparrow$} & \textbf{E$\uparrow$} & \textbf{C$\uparrow$} & \textbf{D$\uparrow$} &
\multicolumn{1}{c}{\textbf{WK.a$\uparrow$}} &
\multicolumn{1}{c}{\textbf{Epi.a$\uparrow$}} &
\multicolumn{1}{c}{\textbf{OE$\uparrow$}} &
\multicolumn{1}{c}{\textbf{AE$\uparrow$}} &
\multicolumn{1}{c}{\textbf{AD$\uparrow$}} &
\multicolumn{1}{c}{\textbf{AT$\downarrow$}} &
\multicolumn{1}{c}{\textbf{IA$\downarrow$}} \\
\midrule
Fixed Size & GPT-5 & STM & 0   & 0 & 0 & 2 & 0 & 2 & 0.322 & 0.208 & 2 / 83  & 10 / 22 & 0.414 & 1062.4  & 0.02 \\
Fixed Size & GPT-5 & STM & 1   & 0 & 0 & 1 & 0 & 3 & 0.374 & 0.369 & 11 / 83 & 13 / 22 & 0.529 & 1491.6  & 0.01 \\
Fixed Size & GPT-5 & STM & 5   & 1 & 2 & 2 & 0 & 5 & 0.409 & 0.587 & 9 / 83  & 16 / 22 & 0.628 & 2362.8  & 0.01 \\
Fixed Size & GPT-5 & STM & 10  & 1 & 2 & 2 & 2 & 6 & 0.426 & 0.366 & 17 / 83 & 17 / 22 & 0.743 & 4404.7  & 0.01 \\
Fixed Size & GPT-5 & STM & 20  & 1 & 3 & 7 & 5 & 6 & 0.496 & 0.376 & 18 / 83 & 19 / 22 & 0.758 & 6817.9  & 0.02 \\
Fixed Size & GPT-5 & STM & 500 & 3 & 2 & 4 & 6 & 1 & 0.713 & 0.918 & 18 / 83 & 17 / 22 & 0.631 & 51856.8 & 0.00 \\
\bottomrule
\end{tabular}%
}
\label{tab:stm_size_analysis}
\end{table}

\begin{table}[H]
\centering
\small
\setlength{\tabcolsep}{3pt}
\caption{\textbf{Full results of three runs of the GPT-5 RAG baseline on Remnant.} Q is main quest reward. SQ, E, C, and D are supplementary rewards for side quests, area exploration, crafting, and defeating, respectively. WK.a and Epi.a are World Knowledge QA and Episodic Memory QA accuracy after gameplay. OE and AE report object and action exploration counts; AD is action diversity, AT is average tokens per step, and IA is the invalid-action rate. Performance is ranked by main quest reward, with supplementary reward breaking ties.}
\label{tab:multiple_runs_variance}
\begin{tabular}{crrrrrrrrrrrr}
\toprule
Run & Q$\uparrow$ & SQ$\uparrow$ & E$\uparrow$ & C$\uparrow$ & D$\uparrow$ & WK.a$\uparrow$ & Epi.a$\uparrow$ & OE$\uparrow$ & AE$\uparrow$ & AD$\uparrow$ & AT$\downarrow$ & IA$\downarrow$ \\
\midrule
0 & 1 & 0 & 2 & 0 & 2 & 0.487 & 0.709 & 6/83 & 12/22 & 0.311 & 2051.3 & 0.00 \\
1 & 1 & 0 & 2 & 0 & 3 & 0.452 & 0.614 & 6/83 & 12/22 & 0.296 & 2183.0 & 0.00 \\
2 & 1 & 2 & 2 & 0 & 0 & 0.513 & 0.840 & 3/83 & 12/22 & 0.171 & 2102.4 & 0.00 \\
\bottomrule
\end{tabular}
\end{table}

%% file: sections/additional_results.tex
\section{Full results of the multi-game evaluation}
\label{app:multi_game_results}

We evaluate Long Context, Fixed Size STM, and RAG+STM with GPT-5, and SFT+STM with Qwen3-4B. Tables~\ref{tab:four_game_results} and~\ref{tab:mark_gpt5_results} give the detailed results underlying Table~\ref{tab:multi_game_results}. The four additional games were generated before this evaluation. We conducted 16 runs on these games and three additional GPT-5 runs on Mark. Each run has a budget of up to 500 steps. GPT-5 Long Context completed Mark in 192 steps. The Qwen3-4B SFT+STM result on Mark is already reported in Table~\ref{tab:results_exp2} and is included below for reference.

\begin{table}[ht]
\centering
\scriptsize
\setlength{\tabcolsep}{4pt}
\caption{\textbf{Full results on four additional games.} All runs last 500 steps. $Q$ is main quest reward. $SQ$, $E$, $C$, and $D$ are rewards for side quests, area exploration, crafting, and defeating. $OE$ and $AE$ give object and action coverage with the totals recorded in the evaluation world definition and action set. $AD$ is action diversity, $AT$ is average tokens per step, and $IA$ is the invalid action rate in percent. Highest main quest rewards are bold.}
\label{tab:four_game_results}
\resizebox{\textwidth}{!}{%
\begin{tabular}{llrrrrrllrrr}
\toprule
Category & Method & $Q\uparrow$ & $SQ\uparrow$ & $E\uparrow$ & $C\uparrow$ & $D\uparrow$ & $OE\uparrow$ & $AE\uparrow$ & $AD\uparrow$ & $AT\downarrow$ & $IA(\%)\downarrow$ \\
\midrule
\multicolumn{12}{l}{\textbf{Metropolis}} \\
Long Context & Baseline & \textbf{9} & 4 & 6 & 3 & 1 & 13/47 & 17/20 & 0.748 & 62,587 & 0.0 \\
Fixed Size & STM & 4 & 4 & 5 & 0 & 0 & 15/47 & 18/20 & 0.758 & 2,633 & 3.8 \\
RAG & STM & 8 & 7 & 6 & 2 & 2 & 12/47 & 15/20 & 0.725 & 3,644 & 1.6 \\
SFT & STM & 7 & 2 & 4 & 0 & 0 & 7/47 & 17/20 & 0.805 & 2,310 & 9.0 \\
\midrule
\multicolumn{12}{l}{\textbf{Saltglass}} \\
Long Context & Baseline & \textbf{6} & 7 & 7 & 11 & 1 & 20/28 & 19/19 & 0.697 & 55,150 & 0.0 \\
Fixed Size & STM & 3 & 3 & 6 & 1 & 2 & 10/28 & 16/19 & 0.735 & 2,715 & 0.2 \\
RAG & STM & 4 & 3 & 5 & 0 & 2 & 11/28 & 15/19 & 0.699 & 2,902 & 1.0 \\
SFT & STM & 2 & 2 & 5 & 0 & 1 & 7/28 & 18/19 & 0.809 & 2,333 & 7.6 \\
\midrule
\multicolumn{12}{l}{\textbf{Quarantine}} \\
Long Context & Baseline & \textbf{13} & 1 & 6 & 6 & 9 & 28/79 & 19/23 & 0.649 & 61,425 & 0.0 \\
Fixed Size & STM & 4 & 0 & 2 & 1 & 1 & 15/79 & 15/23 & 0.597 & 2,572 & 0.0 \\
RAG & STM & 4 & 4 & 3 & 3 & 3 & 20/79 & 18/23 & 0.663 & 3,599 & 0.0 \\
SFT & STM & 2 & 2 & 3 & 1 & 2 & 6/79 & 21/23 & 0.852 & 2,622 & 10.8 \\
\midrule
\multicolumn{12}{l}{\textbf{Robot Kingdom}} \\
Long Context & Baseline & \textbf{2} & 3 & 5 & 5 & 0 & 16/96 & 15/23 & 0.604 & 56,425 & 0.0 \\
Fixed Size & STM & 1 & 0 & 7 & 1 & 0 & 12/96 & 14/23 & 0.540 & 2,570 & 0.0 \\
RAG & STM & 1 & 1 & 4 & 0 & 0 & 11/96 & 13/23 & 0.630 & 3,615 & 0.0 \\
SFT & STM & 1 & 1 & 6 & 0 & 1 & 10/96 & 20/23 & 0.737 & 2,651 & 5.2 \\
\bottomrule
\end{tabular}%
}
\end{table}

\begin{table}[ht]
\centering
\scriptsize
\setlength{\tabcolsep}{4pt}
\caption{\textbf{Additional GPT-5 evaluation on Mark.} The SFT+STM row uses Qwen3-4B and reproduces Table~\ref{tab:results_exp2}. Metrics follow Table~\ref{tab:four_game_results}, with invalid action rate expressed in percent. Long Context completes all main quest stages in 192 steps.}
\label{tab:mark_gpt5_results}
\resizebox{\textwidth}{!}{%
\begin{tabular}{llrrrrrrllrrr}
\toprule
Category & Method & Steps & $Q\uparrow$ & $SQ\uparrow$ & $E\uparrow$ & $C\uparrow$ & $D\uparrow$ & $OE\uparrow$ & $AE\uparrow$ & $AD\uparrow$ & $AT\downarrow$ & $IA(\%)\downarrow$ \\
\midrule
Long Context & Baseline & 192 & \textbf{17} & 0 & 10 & 7 & 3 & 17/48 & 14/21 & 0.697 & 22,628 & 0.0 \\
Fixed Size & STM & 500 & 11 & 0 & 7 & 4 & 11 & 16/48 & 18/21 & 0.718 & 2,359 & 0.4 \\
RAG & STM & 500 & 11 & 0 & 13 & 4 & 14 & 16/48 & 14/21 & 0.707 & 3,221 & 0.0 \\
SFT & STM & 500 & 7 & 0 & 10 & 4 & 2 & 9/48 & 18/21 & 0.777 & 2,436.6 & 13.0 \\
\bottomrule
\end{tabular}%
}
\end{table}

\paragraph{Evaluation cost.}
The 16 runs on the four additional games total 8,000 agent steps and approximately 135 million model tokens. The three GPT-5 runs on Mark add 1,192 steps and approximately seven million tokens. These totals are calculated from the reported average tokens per step and actual run lengths, excluding QA evaluation and game generation. This cost motivates evaluation across multiple games complemented by the three-run study.

\paragraph{Game progress and diagnostics.}
Mean quest completion averages the completed fraction of main quest stages over all six games. We rank performance within each game by main quest reward, using supplementary reward to break ties. The diagnostics in these tables measure behavioral coverage and efficiency alongside progress. World Knowledge QA and Episodic Memory QA are reported for the original Remnant and Mark studies in Tables~\ref{tab:results_proprietary_exp1},~\ref{tab:results_open_exp1}, and~\ref{tab:results_exp2}.

%% file: sections/evaluation_validation.tex
\subsection{Evaluation validation and ability mapping}
\label{app:evaluation_validation}

\paragraph{QA construction and correctness.}
Episodic Memory QA is generated entirely from rule-based templates. For World Knowledge QA, LLMs are used only for world-dynamics questions and distractor enrichment. All other questions are rule-based. The LLM is prompted to generate QA pairs directly from the code in a fixed format, as is common in benchmark construction \citep{saad2024ares,biancini2024multiple,guinet2024automated}. We manually checked 30 sampled questions from Remnant across both QA types and found them 100\% correct. Questions are intentionally simple, since these diagnostic tests are designed to isolate semantic world knowledge acquisition and episodic memory recall rather than additional reasoning.

\paragraph{Mapping metrics to the five abilities.}
These diagnostics are not redundant with main quest reward. World Knowledge QA measures acquired game knowledge, Episodic Memory QA measures trajectory memory, and object and action exploration metrics measure coverage. Falling action diversity can reveal collapse, while the invalid action rate measures how well the model follows the instruction to output the correct action format. For SFT, these diagnostics expose degradation missed by quest progress.

Action diversity is normalized action entropy. The curves use windows of 100 actions, and the tables use the full action history after the tutorial.

\begin{table}[ht]
\centering
\small
\caption{\textbf{Mapping evaluation metrics to the five abilities.} Trajectory inspection supplements the quantitative metrics.}
\label{tab:ability_metric_mapping}
\begin{tabularx}{\textwidth}{lX}
\toprule
Ability & Evaluation \\
\midrule
Exploration & Area exploration reward, object and action coverage, and action diversity. \\
Episodic memory & Episodic Memory QA accuracy after gameplay, supported by trajectory inspection. \\
World knowledge acquisition & World Knowledge QA before, during, and after gameplay, including the progressive curve in Figure~\ref{fig:progressive_wk_qa}. \\
Skill learning & Craft and defeat rewards count unique successful crafting and combat outcomes. \\
Planning over long horizons & Main quest progress measures task decomposition and progress tracking. Side quest progress tests switching from the main quest to a side quest and then back. \\
\bottomrule
\end{tabularx}
\end{table}

\paragraph{Inapplicable actions and action space.}
If an action cannot be executed, the engine rejects it and explains why through feedback. The turn still passes, so time and step rules may change the world. We distinguish two cases. Command parsing or format failures, such as unknown verbs, empty inputs, and incorrect argument counts, are included in the invalid action rate. Failures caused by the current state, such as trying to pick up an object that is not in the current area, are not included, but the engine still provides corrective feedback from which the agent can learn world knowledge such as action affordances. Malformed JSON from the LLM is converted to a default \texttt{wait} action before environment dispatch and is not counted as an invalid action by the engine.

Every game shares the base game's 17 core verbs, and generated rules can add new verbs or edit existing ones. Because actions and quests are generated against one structured world, static checks verify that all references are defined consistently. We also run execution tests and one human playtest of 500 steps per experiment. The Mark tester completed all 17 stages, while the Remnant tester completed 9 of 24 within the budget. These results provide empirical measurements of human progress under the fixed budget.

\paragraph{Quest progression.}
The engine awards main quest rewards only after verifying completion of a stage and its prerequisites. We found no stage skipping in the inspected trajectories. All methods in each experiment are evaluated in the same game under the same protocol. Each highlighted curve in Figure~\ref{fig:vis_results_exp1} represents the agent with the highest performance within a paradigm. For example, the vanilla Long Context agent completed three main quest stages, while RAG with short-term memory completed two.

%% file: sections/generator_evaluation.tex
\section{Game generation experiments and qualitative analysis}
\label{app:generator_evaluation}

\subsection{Open-weight models as game generators}
We added experiments using Qwen3.6-35B-A3B as the game generation backbone and compared its outputs with games generated by GPT-5.5. Both models used the same generation specification: a drowned cathedral city where the tide keeps the only records, two places, 14 requested objects, five NPCs, one new action rule, one new step rule, and one quest chapter with branching factor one. We conducted three independent generations per model. Entity generation used direct LLM output, while rule generation and custom quest conditions used Aider.

Table~\ref{tab:generator_comparison} reports generation success, repair iterations, and time. The time comparison also reflects the serving setup, since Qwen ran locally on an H200 while GPT-5.5 used an API. Thus, an open-weight model can run the full pipeline, while a stronger model improves robustness and quality.

\begin{table}[ht]
\centering
\small
\caption{\textbf{Game generation with open-weight and proprietary models.} Execution tests run for 300 steps. The static quest completion check tests required completion stages and consistency of quest references. Passing this check does not establish global solvability.}
\label{tab:generator_comparison}
\begin{tabular}{lrr}
\toprule
Metric & Qwen3.6-35B-A3B & GPT-5.5 \\
\midrule
Pipeline completed & 3/3 & 3/3 \\
Execution smoke test passed & 3/3 & 3/3 \\
Static quest completion check passed & 2/3 & 3/3 \\
Mean rule-generation repair iterations (range) & 2.7 (2--3) & 1.0 (1--1) \\
Median generation time (minutes) & 18 & 5 \\
\bottomrule
\end{tabular}
\end{table}

\subsection{Qualitative analysis of generation failures}
In the open-weight experiment, one Qwen-generated game ran for 300 steps without crashing but failed the static quest completion check because its main quest lacked a valid completion path. Its quest plan omitted the final completion stage. The subsequent Aider fallback initialized the new quest but retained the old template's quest conditions, leaving eight unresolved references. The game could execute, but its main quest could not progress.

In contrast, all three GPT-5.5 games passed both checks and required fewer repair iterations. These results suggest that Qwen can generate executable environments but is less reliable at structured quest planning and maintaining consistency across quest components. The failure also reflects the pipeline's handling of an invalid quest plan, since the fallback did not update the coupled components consistently. The distinction between executable code and a valid quest motivates complementary static checks, execution tests, and human playtesting.

\paragraph{Generation quality and limitations.}
Human playtesting checks reachability within the interaction budget. Runtime and static checks identify implementation failures but do not establish the absence of every exploit or the naturalness of every rule. Unfamiliar vocabulary and causal rules are intentional. The generator prompt can favor interpretable names when desired.